%% file: main.tex
\documentclass[acmtog]{acmart}
\acmSubmissionID{347}

\usepackage{booktabs} %

\usepackage[ruled]{algorithm2e} %

\SetAlFnt{\small}
\SetAlCapFnt{\small}
\SetAlCapNameFnt{\small}
\SetAlCapHSkip{0pt}

\acmJournal{TOG}
\acmYear{2024} \acmVolume{43} \acmNumber{4} \acmArticle{52} \acmMonth{7}\acmDOI{10.1145/3658157}

\setcopyright{rightsretained}

\input{macros}

\begin{document}
\title{Training-Free Consistent Text-to-Image Generation}

\author{Yoad Tewel}
\orcid{0009-0006-8042-0428}
\affiliation{
 \institution{NVIDIA}
 \country{Israel}
 }
 \affiliation{
 \institution{Tel Aviv University}
 \country{Israel}
 }
\email{yoadtewel@mail.tau.ac.il}

\author{Omri Kaduri}
\orcid{0009-0003-2099-6263}
\affiliation{
 \institution{Independent Scientist}
 \country{Israel}
 }
\email{kaduriomri@gmail.com}

\author{Rinon Gal}
\orcid{0000-0003-4875-965X}
\affiliation{
 \institution{NVIDIA}
 \country{Israel}
 }
 \affiliation{
 \institution{Tel Aviv University}
 \country{Israel}
 }
\email{rinong@gmail.com}

\author{Yoni Kasten}
\orcid{0009-0006-9897-0022}
\affiliation{
 \institution{NVIDIA}
 \country{Israel}
 }
\email{ykasten@nvidia.com}

\author{Lior Wolf}
\orcid{0000-0001-5578-8892}
\affiliation{
 \institution{Tel Aviv University}
 \country{Israel}
 }
\email{liorwolf@gmail.com}

\author{Gal Chechik}
\orcid{0000-0001-9164-5303}
\affiliation{
 \institution{NVIDIA}
 \country{Israel}
 }
\email{gal.chechik@gmail.com}

\author{Yuval Atzmon}
\orcid{0000-0003-3817-3698}
\affiliation{
 \institution{NVIDIA}
 \country{Israel}
 }
\email{yatzmon@nvidia.com}

\begin{abstract}

Text-to-image models offer a new level of creative flexibility by allowing users to guide the image generation process through natural language.
However, using these models to consistently portray \textit{the same} subject across diverse prompts remains challenging. Existing approaches fine-tune the model to teach it new words that describe specific user-provided subjects or add image conditioning to the model.
These methods require lengthy per-subject optimization or large-scale pre-training. Moreover, they struggle to align generated images with text prompts and face difficulties in portraying multiple subjects.
Here, we present \textit{\ourmethod}, a \textit{training-free} approach that enables consistent subject generation by sharing the internal activations of the pretrained model.
We introduce a subject-driven shared attention block and correspondence-based feature injection to promote subject consistency between images. Additionally, we develop strategies to encourage layout diversity while maintaining subject consistency.
We compare \ourmethod to a range of baselines, and demonstrate state-of-the-art performance on subject consistency and text alignment, without requiring a single optimization step. Finally, \ourmethod can naturally extend to multi-subject scenarios, and even enable training-free \textit{personalization} for common objects.

Code will be available at our \href{https://consistory-paper.github.io/}{\textcolor{blue}{project page}}.

\end{abstract}


\begin{teaserfigure}
    \setlength{\abovecaptionskip}{2pt}
    \centering
    \includegraphics[width=0.93\textwidth, trim={1.5cm 30.5cm 4.7cm 2cm},clip]{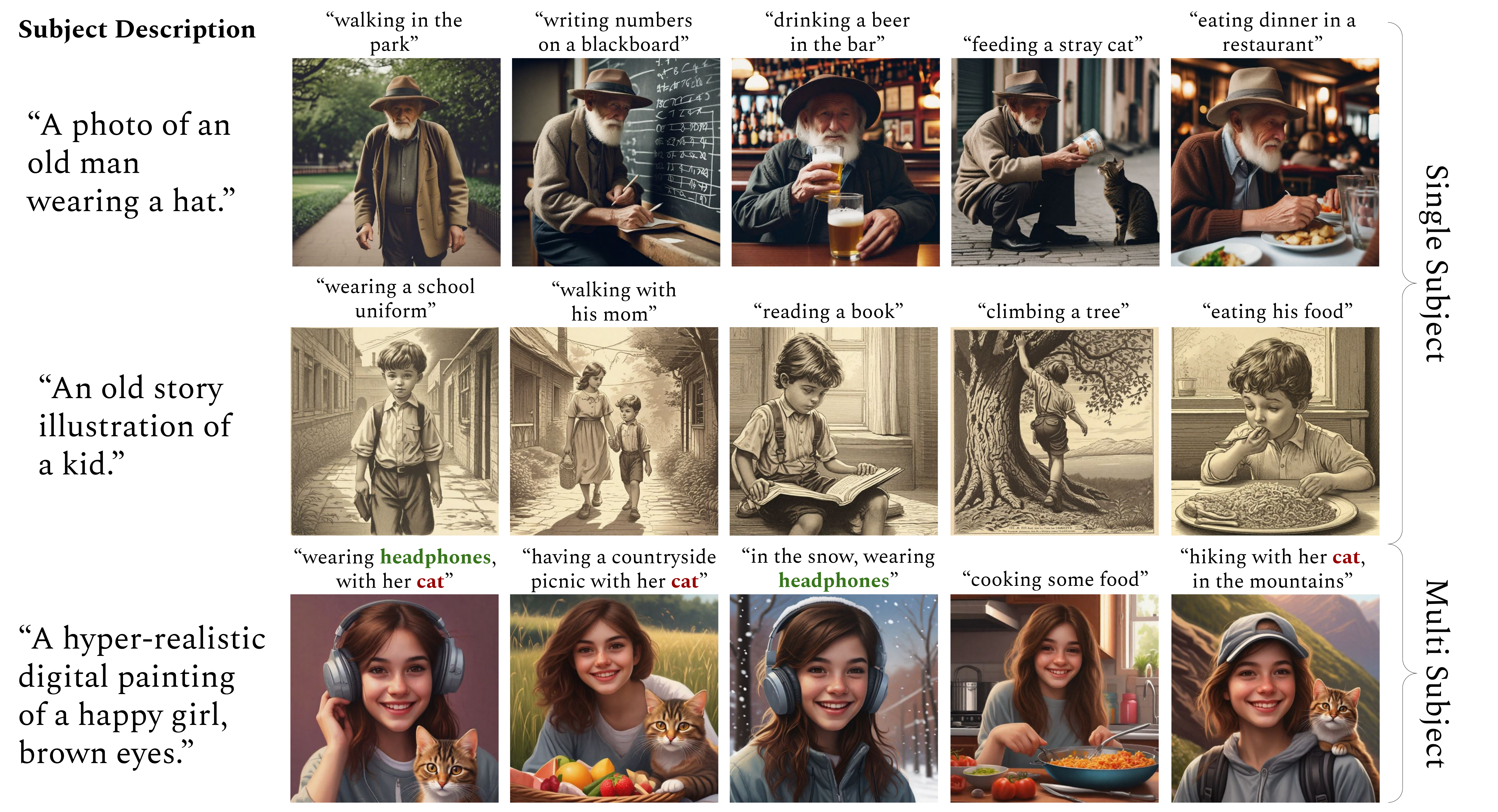} %
    \caption{\textbf{\ourmethod} transforms a set of input prompts with recurring subjects into a series of images that maintain the same subject identity and adhere to the provided text. It can also maintain consistent identities for multiple subjects. Importantly, \ourmethod does not involve any optimization or pre-training.}
    \label{fig:teaser}
\end{teaserfigure}

\maketitle

\section{Introduction}

Large-scale text-to-image (\TTI) diffusion models empower users to create imaginative scenes from text, but their stochastic nature poses challenges when trying to portray \textit{visually consistent} subjects across an array of prompts. Such consistency is crucial for many applications: from illustrating books and stories, through designing virtual assets, to creating graphic novels and synthetic data.

In the field of consistent image generation, current approaches \cite{avrahami2023chosen,liu2023intelligent,feng2023improved,jeong2023zero} predominantly rely on \textit{personalization}, a process where the text-to-image model learns a new word to represent a \textit{specific} subject in a given image set. However, these personalization-based 
methods suffer from several drawbacks: They require per-subject training; they struggle to portray multiple consistent subjects simultaneously in one image; and they can suffer from trade-offs between subject consistency and prompt-alignment. Alternatives, like training image-conditioned diffusion models (\eg using an encoder~\cite{ye2023ip,gal2023encoder,Wei_2023_ICCV}), require significant computational resources, and their extension to multi-object scenes remains unclear. 
A common thread in all these approaches is that they attempt to enforce consistency \textit{a posteriori}. That is, they operate to make generated images consistent with a specific, given target. Such approaches have two drawbacks. They are bound to constrain the model's ``creativity" to the given target image, and they tend to drive the model away from its training distribution. 

We show here that the limitations of a posteriori methods can be avoided, and propose a way to achieve consistency 
in a \textit{zero-shot} manner - without conditioning on existing images. The key idea is to promote cross-frame consistency \textit{a priori} during generation. To achieve this, we leverage the internal feature representations of the diffusion model to align the generated images with each other, without any need to further align them with an external source. In doing so, we can enable on-the-fly consistent generation (\cref{fig:teaser}), without requiring lengthy training or backpropagation, making generation roughly $\times 20$ faster than the current state-of-the-art.

Our approach operates in three steps. First, we localize the subject across a set of noisy generated images. We then encourage subject consistency by allowing each generated image to attend to subject patches in other frames via an extension of the self-attention mechanism. This leads to more consistent subjects across the batch but causes the layout diversity to greatly diminish, as observed in other contexts that use similar extension~\cite{hertz2023StyleAligned}. Our second step is, therefore, to maintain diversity in two ways: by incorporating features from a vanilla, non-consistent sampling step, and by introducing a new inference-time dropout on the shared keys and values.
Finally, we aim to enhance consistency in finer details. To achieve this, we align the self-attention \textit{output} features between corresponding subject pixels across the entire set.

Our full method, which we term \ourmethod, combines these components to enable training-free consistent generation. We compare \ourmethod to prior approaches and demonstrate that by aligning features during the generative process, we not only substantially speed up the process, but also maintain better prompt-alignment. Importantly, our method is trivial to extend to multi-subject scenes, avoiding pitfalls introduced by personalization-based approaches.

Finally, we show that \ourmethod is compatible with existing editing tools like ControlNet~\cite{zhang2023adding}, we introduce methods for re-using the consistent identities, and even apply our ideas to training-free personalization for common object classes, being \textit{the first to show} training-free personalization, with no encoder use.

In summary, this paper makes the following contributions: First, we present a training-free method for achieving subject consistency across varying prompts. Second, we develop new techniques to combat layout collapse in extend-attention applications. Additionally, we share a new benchmark dataset for consistency evaluation.

\section{Related work}

\paragraph{Consistent \TTI generation} is the task of synthesizing a set of images that portray visually consistent subjects. Early works utilized extensive fine-tuning and personalization~\cite{gal2022textual,ruiz2022dreambooth}
to promote consistency. \cite{jeong2023zero} replaces a character face using a personalized model and image editing. \cite{gong2023talecrafter} iteratively generates multi-character images using personalized LoRA models \cite{simoLoRA2023}, and requires pre-training a text-to-layout model. \cite{feng2023improved,liu2023intelligent} involve fine-tuning a \TTI model on storyboard datasets and conditioning it on image frames. This resembles encoder-based personalization methods like IP-Adapter \cite{ye2023ip} and ELITE \cite{Wei_2023_ICCV}. Finally, \cite{avrahami2023chosen} is a concurrent work that trains a personalized LoRA model iteratively by extracting repeated identities from generated image sets. 

\ourmethod does not tune or personalize the pre-trained \TTI model. It seamlessly generates consistent images from text prompts alone.

\paragraph{Attention-based Consistency.}
In the realm of videos, a common practice is to increase temporal consistency by sharing self-attention keys and values~\cite{wu2023tune} across frames. This can be done for generation~\cite{wu2023tune,ceylan2023pix2video,Khachatryan_2023_ICCV} or for video editing~\cite{tokenflow2023,QI_2023_ICCV}. Others use attention keys and values from a source image in order to inject a consistent identity across video frames~\cite{hu2023animate,xu2023magicanimate,chang2023magicdance,tu2023motioneditor}.

When considering images, early works in text-based editing~\cite{hertz2022prompt,tumanyan2023plug,parmar2023zero} proposed to maintain the structure of an image by extracting its attention masks or features, and injecting them into follow-up generations. 

More recent works explored extended-attention mechanisms to maintain consistent appearances when modifying image layouts~\cite{cao_2023_masactrl,mou2023dragondiffusion}, or for training-free appearance-~\cite{alaluf2023crossimage} and style-transfer ~\cite{hertz2023StyleAligned} tasks.

Our method draws on these attention-sharing ideas but applies them to the task of consistent \TTI generation. We do not draw features from existing images or align entire frames, but develop tools to enable subject-level consistency across novel images.

\paragraph{Appearance transfer using dense correspondence maps} has been widely studied. \cite{LiaoDeepImageAnalogy} transfer appearance between images with similar structures using VGG-based maps. \cite{structuralanalogy2020,tumanyan2022splicing} trained generative models to leverage these mappings for image-to-image translation. Recently, diffusion models have been found to establish strong zero-shot correspondence between images \cite{luo2023dhf,hedlin2023unsupervised,zhang2023tale}, enabling applications like instance swapping, image editing, and robust registration. 

Here, we leverage the diffusion-based DIFT maps \cite{tang2023emergent} to share features across multiple images and encourage the generation of subjects with consistent appearance. This aligns features throughout the denoising process rather than doing appearance transfer as a post-hoc step.

\begin{figure*}[!ht]
    \setlength{\abovecaptionskip}{4pt}
    \setlength{\belowcaptionskip}{0pt}
\centering
\includegraphics[width=0.96\textwidth, trim={0.cm 0.5cm 0.cm 0.cm},clip]{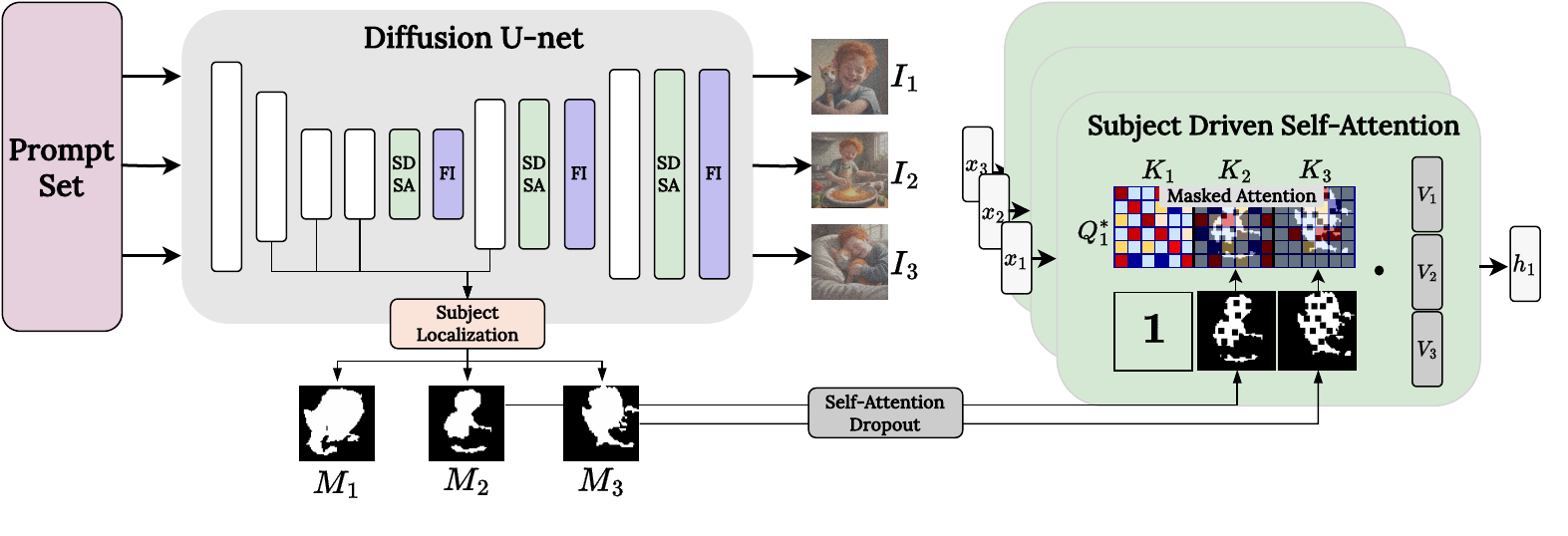} %
    \caption{
    \textbf{Architecture outline (left):} Given a set of prompts, at every generation step we localize the subject in each generated image $I_i$. We utilize the cross-attention maps up to the current generation step, to create subject masks $M_i$. Then, we replace the standard self-attention layers in the U-net decoder with Subject Driven Self-Attention layers that share information between subject instances. We also add Feature Injection for additional refinement. \textbf{Subject Driven Self-Attention (right):} We extend the self-attention layer so the Query from generated image $I_i$ will also have access to the Keys from all other images in the batch ($I_j$, where $j \neq i$), restricted by their subject masks $M_j$.
    To enrich diversity we: (1) Weaken the SDSA via dropout and (2) Blend Query features with vanilla Query features from a non-consistent sampling step, yielding $Q_1^*$.%
    }
    \label{fig_architecture}
\end{figure*}

\section{Preliminaries: Self-Attention in \TTI models}
\label{sec_notation}

Our method manipulates self-attention in \TTI diffusion models. We start by outlining its mechanism and introducing key notations.

A self-attention layer receives a series of tokens, each of which contains features describing a single image \textit{patch}. Each such token undergoes linear projections through three self-attention matrices: $\mW_K$, $\mW_V$ and $\mW_Q$. The results of these projections are known as ``Keys'', ``Values'' and ``Queries'', respectively.

More concretely, consider the $i^{th}$ image entry in the generated batch. Let $x_i \in \R^{P\times d}$ be a sequence of $P$ input token vectors with feature dimension $d$. We define $K_i=x_i \cdot \mW_K$, $V_i=x_i \cdot \mW_V$, $Q_i=x_i \cdot \mW_Q$. The self-attention map is then given by:
\begin{equation}
\label{eq_softmax}
    A_i = \textit{softmax}\left({Q_i K_i^\top}/{\sqrt{d_k}}\right) \in \R^{PxP},   
\end{equation}
where $d_k$ is the feature dimension of $\mW_K$, $\mW_Q$ projections.
Intuitively, this map provides a relevancy score between every pair of patches in the image. It is then used to weight how much the ``Value'' features of a given target patch should influence a source patch $h_i = A_i \cdot V_i$,
where $h$ denotes an intermediary, hidden feature set.

These are projected using a fourth, ``output-projection" matrix, $\mW_O$, yielding $x_i^{out} = \mW_O \cdot h_i$, which is then summed with the input features $x_i$ to create the input for the next layer. 

Our method intervenes in this self-attention mechanism by allowing images in  a generated batch to attend to each other, and be influenced by each other's $x^{out}$ activations. %

\section{Method}\label{sec_method}
Our goal is to generate a set of images portraying consistent subjects across an array of prompts. We propose to do so by better aligning the internal activation of the \TTI model during image denoising. Importantly, we aim to enforce consistency exclusively through an inference-based mechanism, without additional training.

Our approach is comprised of three main components. First, we introduce a \textbf{subject-driven self-attention} mechanism (SDSA), aimed at sharing subject-specific information across relevant model activations in the generated image batch. 
Second, we observe that the above component comes at the cost of reducing the variation in the generated layouts. Therefore, we propose strategies for mitigating this form of mode collapse through an attention-dropout mechanism, and by blending query features obtained from a vanilla, non-consistent, sampling step. 
Third, we incorporate a \textbf{feature injection} mechanism to further refine the results. There, we map features from one generated image to another based on a cross-image dense-correspondence map derived from the diffusion features.
Below, we outline each of these components in detail.

\begin{figure}[t]
    \setlength{\abovecaptionskip}{4pt}
    \setlength{\belowcaptionskip}{0pt}
    \centering
    \includegraphics[width=0.9\columnwidth, trim={0.2cm 0cm 0cm 0cm},clip]{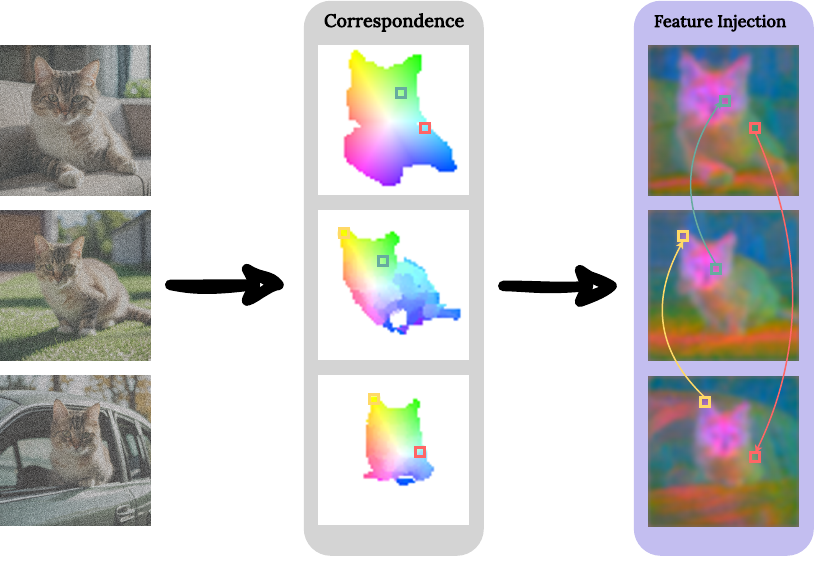} %
    \caption{ \textbf{Feature Injection:} To further refine the subject's identity across images, we introduce a mechanism for blending features within the batch. We extract a patch correspondence map between each pair of images (Middle), and then inject features between images based on that map (Right).}
    \label{fig_feature_injection}
\end{figure}

\subsection{Subject-driven self-attention}

Consider a simple idea for promoting consistency: expanding the self-attention, so that queries from one image can also attend to keys and values from other images in the batch.
This enables repeated objects to naturally attend to each other, thus sharing visual features across images.
This idea is often used in video generation and editing works~\cite{wu2023tune}, leading to increased consistency across frames. However, generated videos differ from our scenario. First, they are created with a single prompt that is shared across frames. Second, they typically require little variation in backgrounds or layout from one frame to the next.
In contrast, we want each frame to follow a unique prompt, and we want to maintain diversity in backgrounds and layouts. \Naive{ly} employing these video-based mechanisms leads to uniform backgrounds and drastically reduced alignment with each image's prompt - in line with expectations for a single video scene, but in direct conflict with our objectives.

One way to tackle these limitations is by reducing the amount of information being shared at background patches. As we are only concerned about sharing subject appearance, we mask the expanded self-attention, so that queries from one image can only match keys and values from the same image, or from regions containing the subject in other images. This way, features for repeated subject elements can be shared, while background features remain separate. 

To this end, we employ a similar approach to prior art~\cite{cao_2023_masactrl,chefer2023attend} and identify noisy latent patches that are likely to contain the subject using cross-attention features. Specifically, we average and threshold the cross-attention maps related to the subject token across diffusion steps and layers to create subject-specific masks (details in Appendix \ref{supp_implementation_details}). With these masks, we propose Subject-Driven Self-Attention (SDSA) where attention is masked so each image can only attend to its own patches or the subject patches within the batch (See \cref{fig_architecture}).
\begin{align}
    K^{+} &= [K_1 \oplus K_2 \oplus \ldots \oplus K_N] \in \R^{N\cdot P \times d_k} \nonumber \\
    V^{+} &= [V_1 \oplus V_2 \oplus \ldots \oplus V_N] \in \R^{N\cdot P \times d_v}   \nonumber \\
    M^{+}_{i} &= [M_1 \ldots M_{i-1} \oplus \mathds{1} \oplus M_{i+1} \ldots M_N] \\
    A^{+}_i &= \textit{softmax}\left({Q_i K^{+\top}}/{\sqrt{d_k}} + \log M^{+}_{i} \right)  \in \R^{P \times N\cdot P} \nonumber \\
    h_i &= A^{+}_i \cdot V^{+} \in \R^{P \times d_v}.
\end{align}

Here, $M_{i}$ is the subject mask for the $i^{th}$ entry in the batch, and $\oplus$ indicates matrix concatenation. We use standard attention masking, which null-out softmax's logits by assigning their scores to $-\infty$ according to the mask. Note that the Query tensors remain unaltered, and that the concatenated mask $M^{+}_i$ is set to be an array of $1$'s for patch indices that belong to the $i^{th}$ image itself.

\subsection{Enriching layout diversity} \label{sec_diversity}
The use of SDSA restores prompt alignment and avoids background collapse. However, we observe that it can still lead to excessive similarity between image layouts. For example, subjects will typically be generated in similar locations and poses.

To improve the diversity of our results, we propose two strategies: first, incorporating features from a vanilla, non-consistent sampling step; and second, further \textit{weakening} the subject-driven shared attention through a dropout mechanism. 

\paragraph{Using Vanilla Query Features.} Recent work~\cite{alaluf2023crossimage}, demonstrated that one can use diffusion models to combine the appearance of one image with the structure of another. They do so by injecting self-attention Keys and Values from the appearance image, and Queries from the structure image. Inspired by this, we aim to enhance pose variation by aligning more closely with a structure predicted by a more diverse vanilla forward pass (\ie without our modifications). We focus on the early steps of the diffusion process, which have been shown to primarily control layout \cite{balaji2022ediffi,patashnik2023localizing}, and apply the following query-blending mechanism: Let ${z_t}$ be the noisy latents at step $t$. We first apply a vanilla denoising step to ${z_t}$, without SDSA, and cache the self-attention queries generated by the diffusion network: ${Q^{vanilla}_t}$. Then, we denoise the same latents ${z_t}$ again, this time using SDSA. During this second pass, for all SDSA layers, we linearly interpolate the generated queries towards the vanilla queries, resulting in: 
\begin{equation}
    Q_t^* = (1 - \nu_t)Q^{SDSA}_t + \nu_t Q^{vanilla}_t,
\end{equation}
where $v_t$ is a linearly decaying blending parameter (see Appendix). %

\paragraph{Self-Attention Dropout} Our second strategy to enhance layout variation involves weakening SDSA using a dropout mechanism. Specifically, at each denoising step, we randomly nullify a subset of patches from $M_i$ by setting them to $0$. This weakens the attention sharing between different images and subsequently promotes richer layout variations. Notably, by adjusting the dropout probability, we can regulate the strength of consistency, and strike a balance between visual consistency and layout variations.

Through these two mechanisms, we aim to tackle two aspects of the layout-collapse problem: Query-feature blending allows to retain aspects of diversity from the non-consistent sampling, while attention-dropout encourages the model to rely less on the shared keys and values, avoiding over-consistency. By mixing them, we achieve increased diversity without significant harm to consistency.

\begin{figure*}[!ht]
    \setlength{\abovecaptionskip}{4pt}
    \setlength{\belowcaptionskip}{0pt}
    \centering
    \includegraphics[width=0.88\textwidth, trim={10.0cm 61.0cm 54.0cm 4.cm},clip]{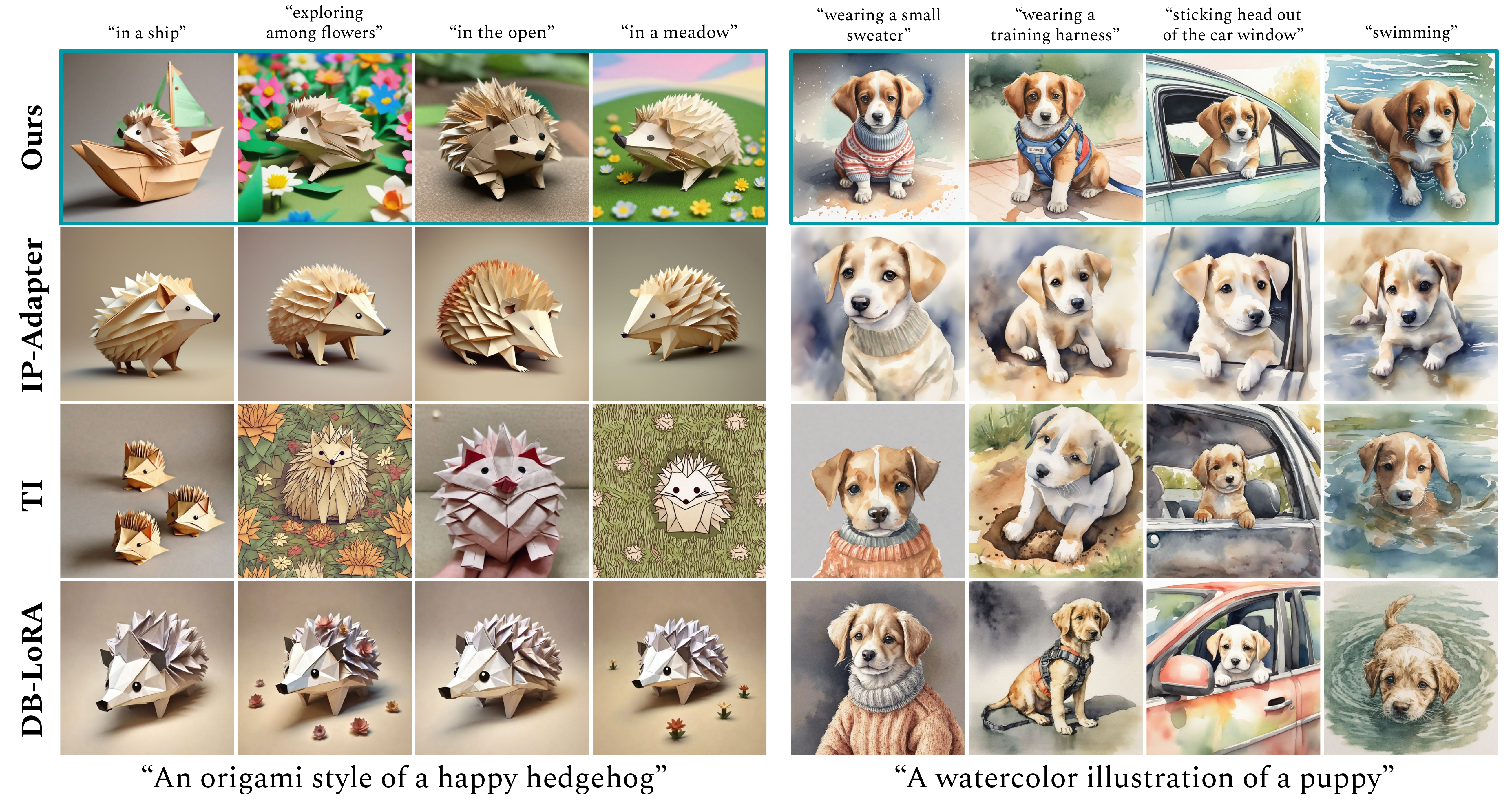} %
    \caption{
    \textbf{Qualitative Results}
    We evaluated our method against IP-Adapter, TI, and DB-LORA.
    Some methods failed to maintain consistency (TI), or follow the prompt (IP-Adapter). Other methods alternated between keeping consistency or following text, but not both (DB-LoRA). Our method successfully followed the prompt while maintaining consistency. Additional results are shown at \figref{fig:fig_extra_qualitative_baselines}}
    \label{fig_single}
\end{figure*}

\subsection{Feature injection}
The shared attention mechanism notably improves subject consistency but may struggle with fine visual features, which may hurt the subject's identity. Hence, we propose to further improve consistency through a novel cross-image \textbf{Feature Injection} mechanism.

Here, we aim to improve the similarity of features from corresponding regions (\eg the left eye) across different images in the batch. Specifically, we find that substantial texture information is contained in the self-attention output features,
$\vx^{out}$, and aligning these features between matching areas can enhance consistency.

To align these features, we first build a patch correspondence map between every pair of images $I_{t}$ and $I_{s}$ in the batch, using DIFT~\cite{tang2023emergent} features $D_t$ and $D_s$ (See Appendix). We denote the correspondence map by $C_{t \rightarrow s}$.  Intuitively, when applied on patch index $p$ from $I_t$, $C_{t \rightarrow s}[p]$ yields the most similar patch in $I_{s}$, as illustrated in \cref{fig_feature_injection}.

Then, to promote feature similarity, we can blend corresponding features based on this mapping. We extend this idea to a \textit{many-to-one} scenario, where each image $I_{t}$ is blended with the other images in the batch. For each patch index $p$ in image $I_t$, we compare its corresponding patches in all other images and select the one with the highest cosine similarity in the DIFT feature space. Formally:
\begin{equation}
\text{src}(p) = \argmax_{s \neq t} \textit{similarity}(D_t[p], D_s[C_{t \rightarrow s}[p]]),
\end{equation}
where $\text{src}(p)$ is the ``best'' source patch for the target patch $p$, and $\textit{similarity}$ is the cosine similarity score.

Finally, we blend the self-attention output layer features of the target image $\vx_t^{out}$, and its corresponding source patches, $\vx_s^{out}$.

\begin{align}
\hat{\vx_t}^{out} &= (1-\alpha) \cdot \vx_t^{out}  + \alpha \cdot \text{src}(\vx_t^{out}),
\end{align}
where $\alpha$ is a blending parameter, and $\text{src}(\vx_t^{out}) \in \R^{P \times d}$ is the tensor obtained by pooling the corresponding features for each patch $p$ in $\vx_t^{out}$ from the associated patch $src(p)$.

In practice, to enforce consistency between appearances of the same subject, without affecting the background, we exclusively apply the feature injection according to the subject masks $M_i$. Additionally, we apply a threshold to inject features only between patches with high enough similarity in the DIFT space (see Appendix).
This approach ensures that features contributing to the appearance of the subject are collectively drawn from all source images, promoting a more comprehensive and representative synthesis.

\subsection{Anchor images and reusable subjects}
As an additional optimization, we can reduce the computational complexity of our approach by designating a subset of generated images as \textbf{``anchor images''}. Rather than sharing keys and values across all generated images during SDSA steps, we allow the images to only observe keys and values derived from the anchors. Similarly, for feature injection, we only consider the anchors as valid feature sources. Note that the anchors can still observe each other during generation, but they too do not observe features from non-anchor images. We find that for most cases, two anchors are sufficient.

This offers several benefits: First, it allows for faster inference and reduced VRAM requirements, because it restricts the size of extended attention. Second, it can improve generation quality in large batches, where we notice it can reduce visual artifacts. Most importantly, we can now reuse the same subjects in novel scenes by creating a new batch where the same prompts and seeds are used to re-create the anchor images, but the non-anchor prompts have changed. Through this mechanism, our approach can yield \textbf{reusable subjects}, and unlimited consistent image generation. %

\subsection{Multi-subject consistent generation}
Personalization-based approaches struggle in maintaining consistency over \textit{multiple} subjects within a single image~\cite{tewel2023key,kumari2022customdiffusion,po2023orthogonal,gu2023mix}. However, with \ourmethod, multi-subject consistent generation is possible in a simple, straightforward manner, by simply taking a \textit{union} of the subject masks. When the subjects are semantically different, information leakage between them is not a concern. This is due to the exponential form of the attention softmax, which acts as a gate that suppresses information leakage between unrelated subjects. Similarly, thresholding the correspondence map during feature injection yields a gating effect that safeguards against information leakage.
Additional details such as hyperparameter choices are in the supplementary.

\begin{figure}[t]
    \setlength{\abovecaptionskip}{4pt}
    \setlength{\belowcaptionskip}{0pt}
    \centering
    \includegraphics[width=1.\columnwidth, trim={1cm 85cm 102cm 1cm},clip]{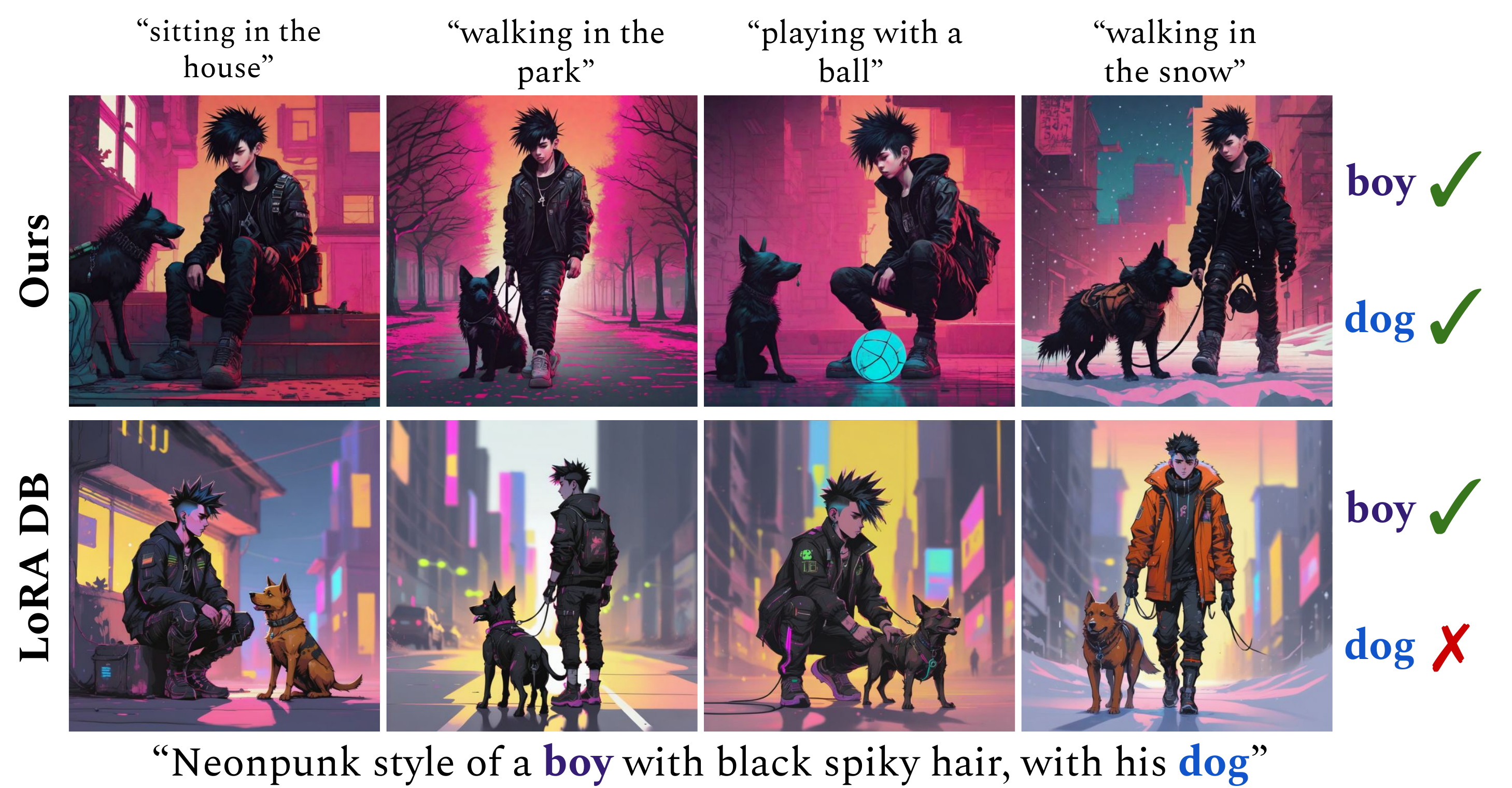} %
    \caption{ \textbf{Multiple Subjects:} \ourmethod generates multiple consistent subjects, while other methods often neglect at least one subject.}
    \label{fig_multi}
\end{figure}

\section{Experiments}
We begin our evaluation by comparing \ourmethod with a range of prior and concurrent baselines. We open with a qualitative comparison, showing that our method can achieve improved subject-consistency and higher prompt-alignment when compared to the state-of-the-art. Next, we evaluate our method quantitatively, including a large-scale user study which demonstrates that users typically favor our results.
Moving on, we conduct an ablation study to highlight the contribution and effect of each component in our method. Finally, we conclude with a set of extended applications, showing that our method is compatible with existing tools such as ControlNet~\cite{zhang2023adding}, and it can even be used to enable training-free personalization for common object classes.

\subsection{Evaluation baselines}
We compare our method to three classes of baselines:
\textbf{(1)} The baseline SDXL model, without adaptations.
\textbf{(2)} Optimization-based personalization approaches that teach the model about a new subject by fine-tuning parts of the model: Textual Inversion (TI) \cite{gal2022textual} fine-tunes the text encoder's word embeddings. DreamBooth-LoRA (DB-LORA) ~\cite{simoLoRA2023} tunes the diffusion U-Net using Low Rank Adaptation~\cite{Hu2021LoRALA}.
\textbf{(3)} Encoder-based approaches that take a single image as input and output a conditioning code to the diffusion model: IP-Adapter \cite{ye2023ip} and ELITE \cite{Wei_2023_ICCV}.
For the personalization and encoder baselines, we first generate a single image of a target subject using a prompt describing the subject, then use it to personalize the diffusion model. All methods except ELITE are based on a pre-trained SDXL model.

For \ourmethod, we use two anchor images and $0.5$ dropout. For the automated metric, we also use lower dropout values.

\subsection{Qualitative Results}
\begin{figure}[!htb]
    \centering
    \includegraphics[width=\columnwidth, trim={1cm 73.5cm 99cm 0cm},clip]{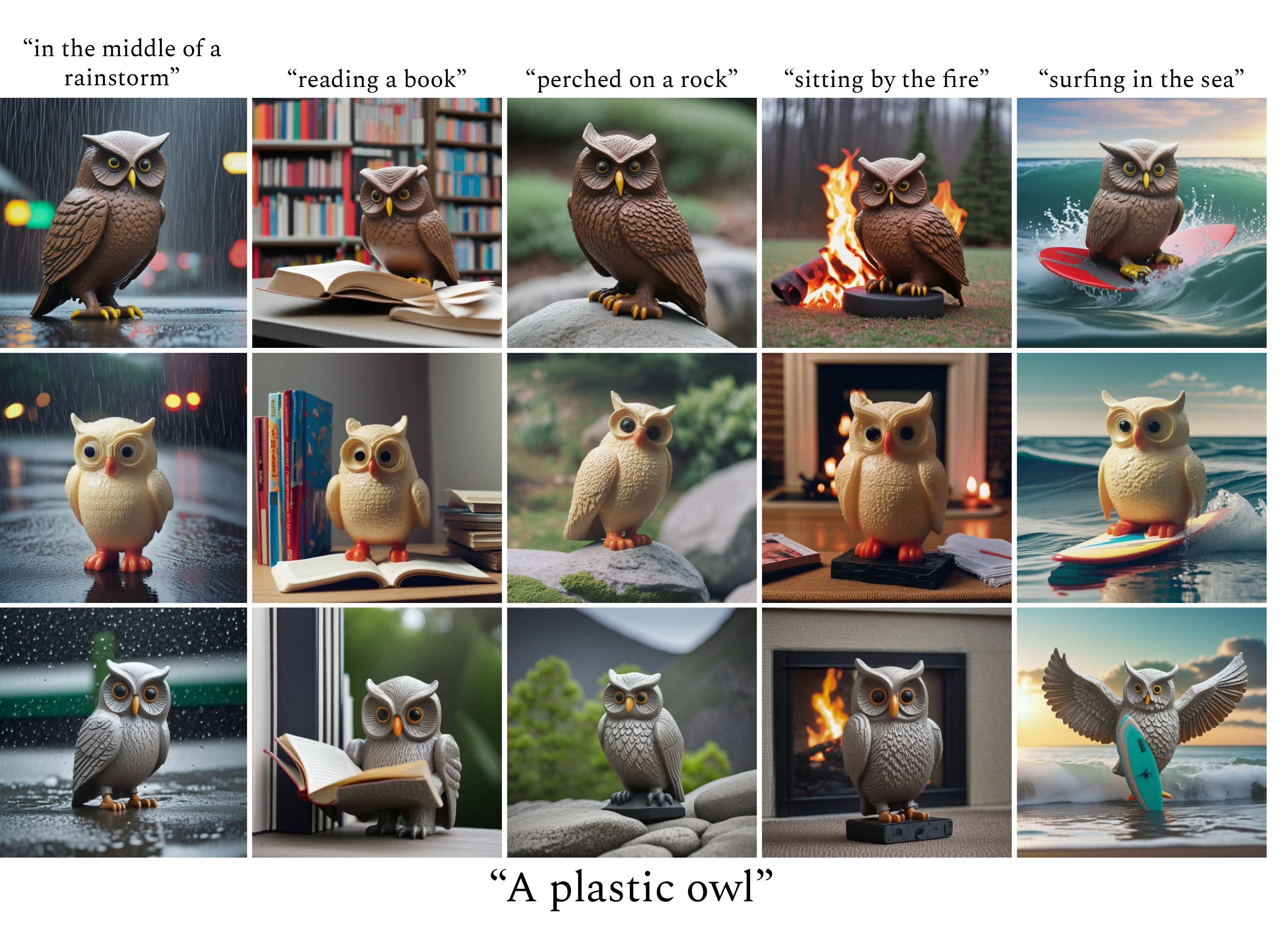} %
    \caption{ \textbf{Seed Variation.} Given different starting noise, \ourmethod generates different consistent set of images.}
    \label{fig:seed_variation}
\end{figure}

\begin{figure}[!hbt]
    \setlength{\abovecaptionskip}{4pt}
    \setlength{\belowcaptionskip}{0pt}
    \centering
    \includegraphics[width=0.98\columnwidth, trim={1cm 95cm 100cm 1cm},clip]{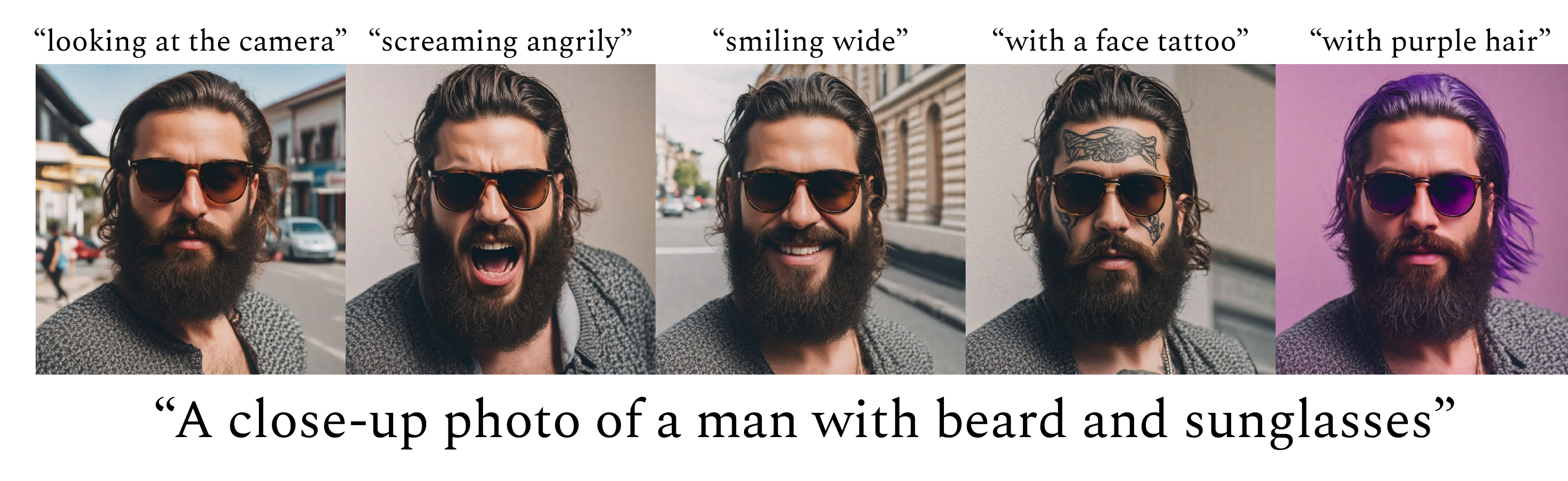} %
    \caption{\textbf{Facial expression variation: \ourmethod is able to generate images of the same identity while facial attributes.}}
    \label{fig:faces_variation}
\end{figure}

In \cref{fig_single} and \cref{fig:fig_extra_qualitative_baselines}, we show qualitative comparisons. Our method can achieve a high degree of subject consistency, while better adhering to the text prompts. Tuning-based personalization approaches tend to either overfit the single training image, producing no variations, or underfit and fail to maintain consistency. IP-Adapter similarly struggles to match complex prompts, particularly when styles are involved. Our method can successfully achieve both subject consistency and text-alignment. Additionally, in \cref{fig:seed_variation} we show that with different initial noise inputs, our method can generate varied sets of consistent images. In \cref{fig:faces_variation} we focus on photo-realistic face imagery and demonstrate that we can maintain a consistent identity while varying attributes such as expression, hair color, or tattoos.

\paragraph{Multi-Subject Generation}
In \cref{fig:teaser} (bottom) we demonstrate that \ourmethod can create scenes with multiple consistent subjects. In \cref{fig_multi} we further compare our method to LORA-DB. Notably, LORA-DB tends to neglect the consistency of one or even both subjects. This pitfall is common when combining personalization approaches, as they learn each subject in isolation. In contrast, our method simply builds on the diffusion model's inherent compositional ability. In Figures \ref{fig:fig_extra_qualitative_vs_dblora}, \ref{fig:chosen_one_multi} we provide additional comparisons. %

\begin{figure*}[!ht]
    \centering
    \includegraphics[width=0.98\textwidth, trim={1.0cm 15.0cm 70.0cm 0.cm},clip]{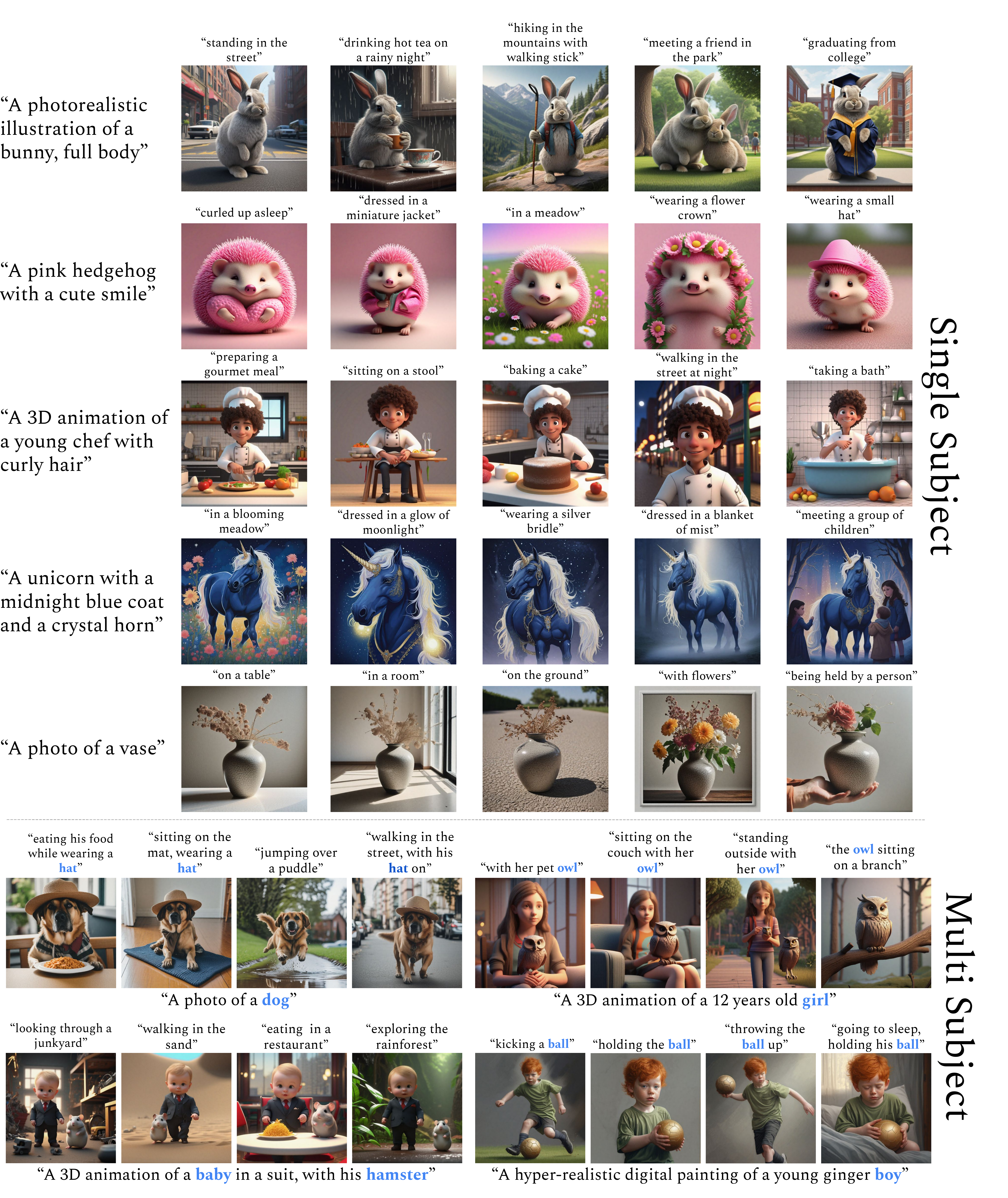} %
    \caption{
    \textbf{Additional Qualitative Results} We demonstrate that \ourmethod successfully generates a consistent subject while following the prompt for various subjects (\textbf{Top}). Furthermore, we show that \ourmethod still succeeds even in the case of multiple interacting subjects (\textbf{Bottom}).
    }
    \label{fig:fig_extra_qualitative_ours}
\end{figure*}

\cref{fig:fig_extra_qualitative_ours} provides more results with single and multiple subjects. %

\subsection{Quantitative evaluation}
Next, we perform quantitative evaluations with automated metrics.
First, we use each baseline to generate $100$ image sets, where each set contains $5$ images depicting a shared subject under different prompts. Our evaluation prompts were created using ChatGPT \cite{OpenAI2022ChatGPT}, with the following protocol: Each prompt consisted of three parts: (1) a subject description, \eg, \textit{``A red dragon''} (2) a setting description, \eg, \textit{``blowing bubbles''} or \textit{``in a castle''}, and (3) a style descriptor, \eg, \textit{``Origami style''}. For subject descriptions, we utilized both detailed examples (\textit{``A red dragon''}), and non-detailed ones (\textit{``A dragon''}). For setting descriptions, we asked ChatGPT to provide descriptions that naturally fit the subject. Each of the $100$ sets contains prompts sharing the same subject description and style but with varying setting descriptions. Further details can be found in Appendix \ref{sec_supp_additional_details}.

\begin{figure}[!htb]
    \setlength{\abovecaptionskip}{4pt}
    \setlength{\belowcaptionskip}{0pt}
    \centering
    \includegraphics[width=0.9\columnwidth, trim={0.2cm 0cm 0cm 0cm},clip]{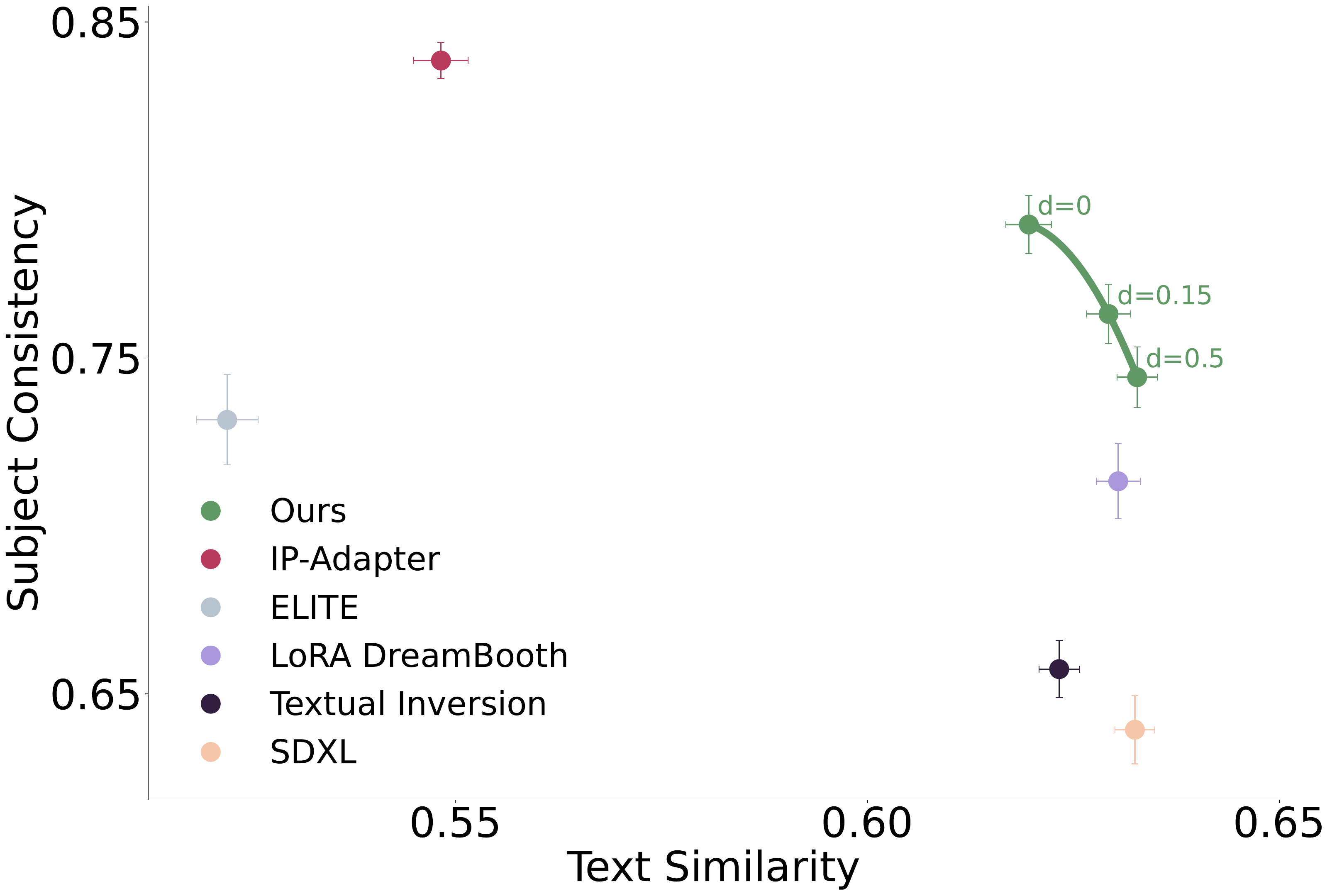} %
    \caption{ \textbf{Subject Consistency VS Textual Similarity:} \ourmethod (green) achieves the optimal balance between Subject Consistency and Textual Similarity. Encoder-based methods such as ELITE and IP-Adapter often overfit to visual appearance, while optimization-based methods such as LoRA-DB and TI do not exhibit high subject consistency as in our method. $d$ denotes different self-attention dropout values. Error bars are S.E.M.}
    \label{fig_main_pareto}
\end{figure}

\begin{figure}[!htb]
    \centering
    \setlength{\abovecaptionskip}{2pt}
    \setlength{\belowcaptionskip}{-2pt}
    \includegraphics[width=0.9\columnwidth, trim={0.2cm 0cm 0cm 0cm},clip]{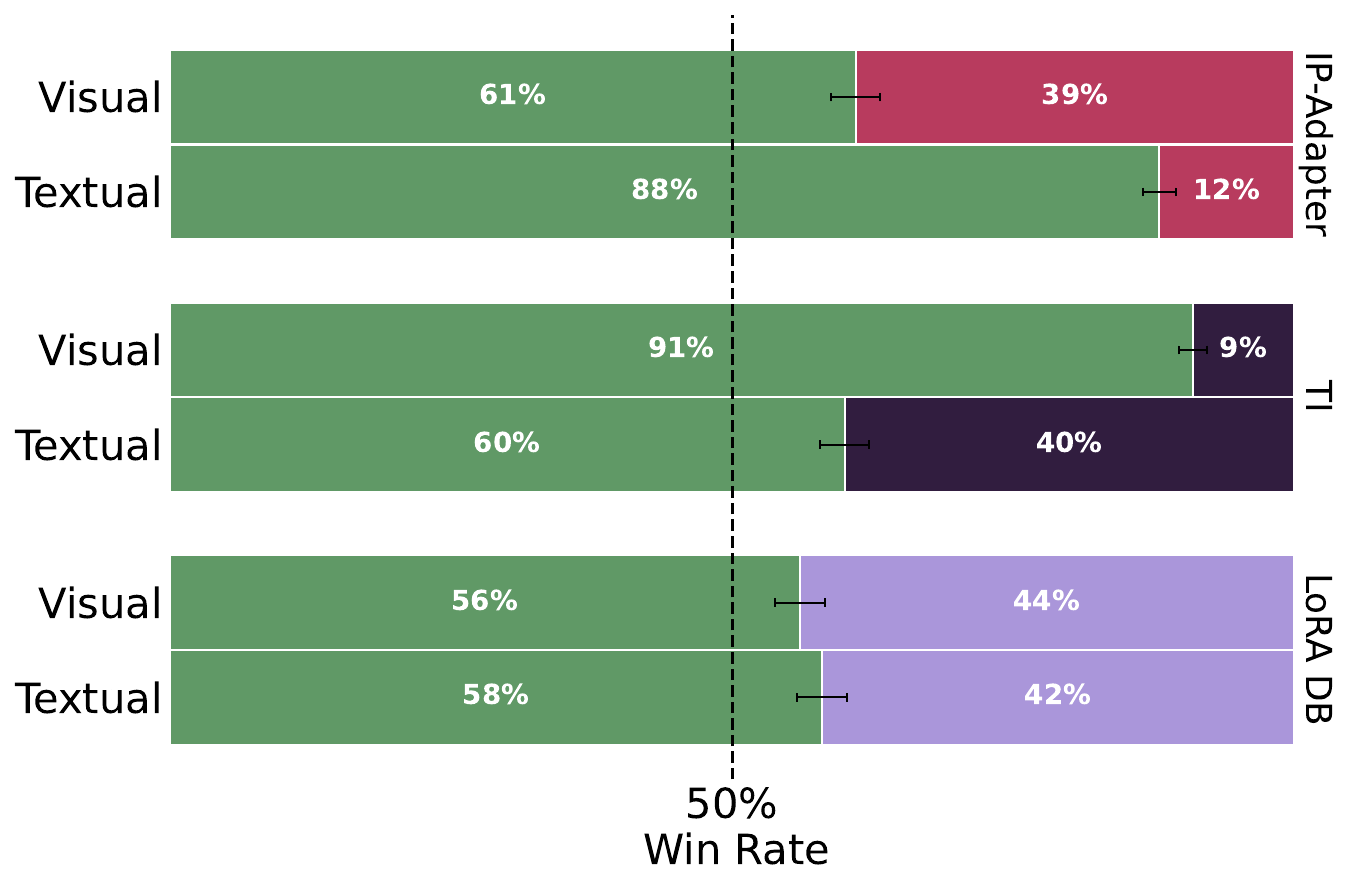} %
    \caption{ \textbf{User Study} results indicate a notable preference among participants for our generated images both in regards to Subject Consistency (\textit{Visual}) and Textual Similarity (\textit{Textual}).}
    \label{fig:userstudy}
\end{figure}

\begin{figure*}[!t]
    \setlength{\abovecaptionskip}{4pt}
    \setlength{\belowcaptionskip}{0pt}
    \centering
    \includegraphics[width=0.8\textwidth, trim={4cm 72cm 75cm 1cm},clip]{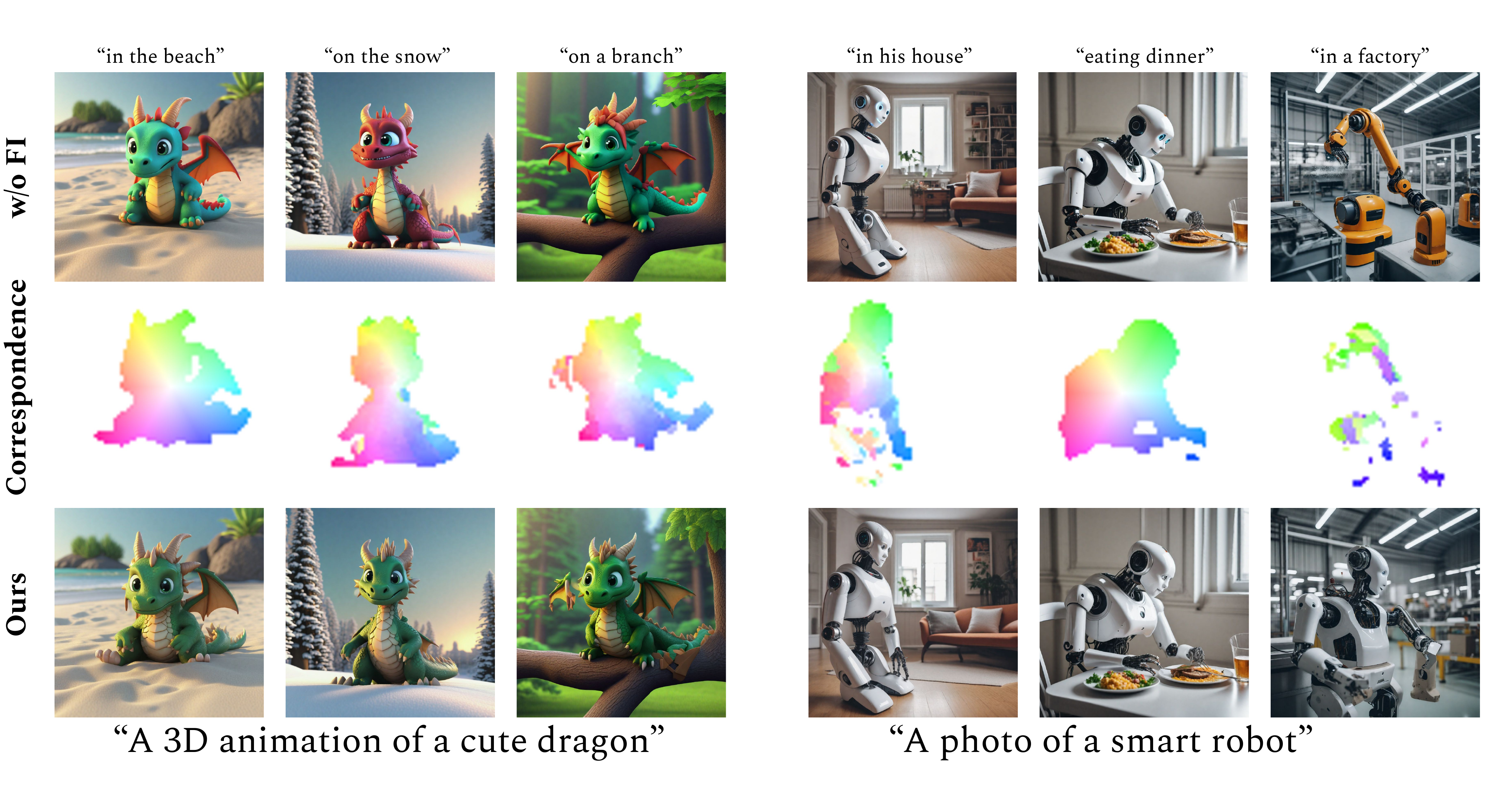} %
    \caption{ \textbf{Additional feature injection results}. Our method can handle scenarios where the estimated correspondence maps are noisy, or even when parts of the objects are obscured. }
    \label{fig_noisy_correspondence}
\end{figure*}

We follow the concurrent work of \citet{avrahami2023chosen} and evaluate the methods on two axes - prompt-alignment, and subject consistency. For prompt-alignment, we use CLIP to measure the similarity between each generated image and its conditioning prompt, and report the average CLIP-score~\cite{hessel-etal-2021-clipscore} over all $500$ generated images. For consistency evaluation, we use DreamSim~\cite{fu2023dreamsim}, which has been shown to better correlate with human judgment of inter-image similarity. We calculate the pair-wise similarity between each pair of images in each of the $100$ sets. To focus on subject consistency, we follow Dreamsim's background removal protocol. We report the average score over these sets. In our results, the error bars represent the Standard Error of the Mean (S.E.M).

The results are provided in \cref{fig_main_pareto}. As can be seen, our method is situated on the Pareto-front. Our standard setup matches SDXL in  text-alignment scores, demonstrating that promoting consistency a-priori can better maintain the model's knowledge. 

However, qualitatively, we observed that the consistency metric tends to bias its scores towards configurations with slight layout changes, rather than assessing consistency in the subject's identity. Therefore, given the limitations of the automated metric, we conducted a large-scale user study.
We concentrated on the most effective techniques, omitting both vanilla SDXL and ELITE. We used the standard two-alternative forced-choice format. Users faced two types of questions: (1) Subject-consistency, where they were shown two sets of 5 images each. They were asked to choose the set that better shows the same subject, ignoring background, pose, and image quality. (2) Text-alignment, where they selected the image that best matched a textual description from two images. We gathered $500$ responses per baseline for each question type, totaling $3,000$ responses.
The results are shown in \cref{fig:userstudy}. Despite being a training-free approach, \ourmethod outperforms the baselines, both regarding textual alignment and subject consistency.

\paragraph{Runtime comparison}
We conducted a runtime analysis of the main methods, focusing on their time-to-consistent-subject (TTCS) using an H100 GPU. Our method, \ourmethod, achieved the fastest TTCS result, at 32 seconds for generating two anchors and an image based on a new prompt. This is $\times 25$ faster than the SoTA approach by \citet{avrahami2023chosen} which is estimated at 13 minutes on an H100 GPU. Moreover, our method is $\times$8-14 faster than LORA-DB (4.5 min.) and TI (7.5 min.). Comparing our approach to encoder-based methods like IP-Adapter is more convoluted. These techniques require \textit{weeks} of pre-training, but once trained, they can generate an anchor and another image based on a new prompt in just 8 seconds.

\subsection{Ablation study}
\label{sec_supp_ablation}

\begin{figure}[t]
    \setlength{\abovecaptionskip}{4pt}
    \setlength{\belowcaptionskip}{0pt}
    \centering
    \includegraphics[width=0.98\columnwidth, trim={0.2cm 69cm 95cm 1cm},clip]{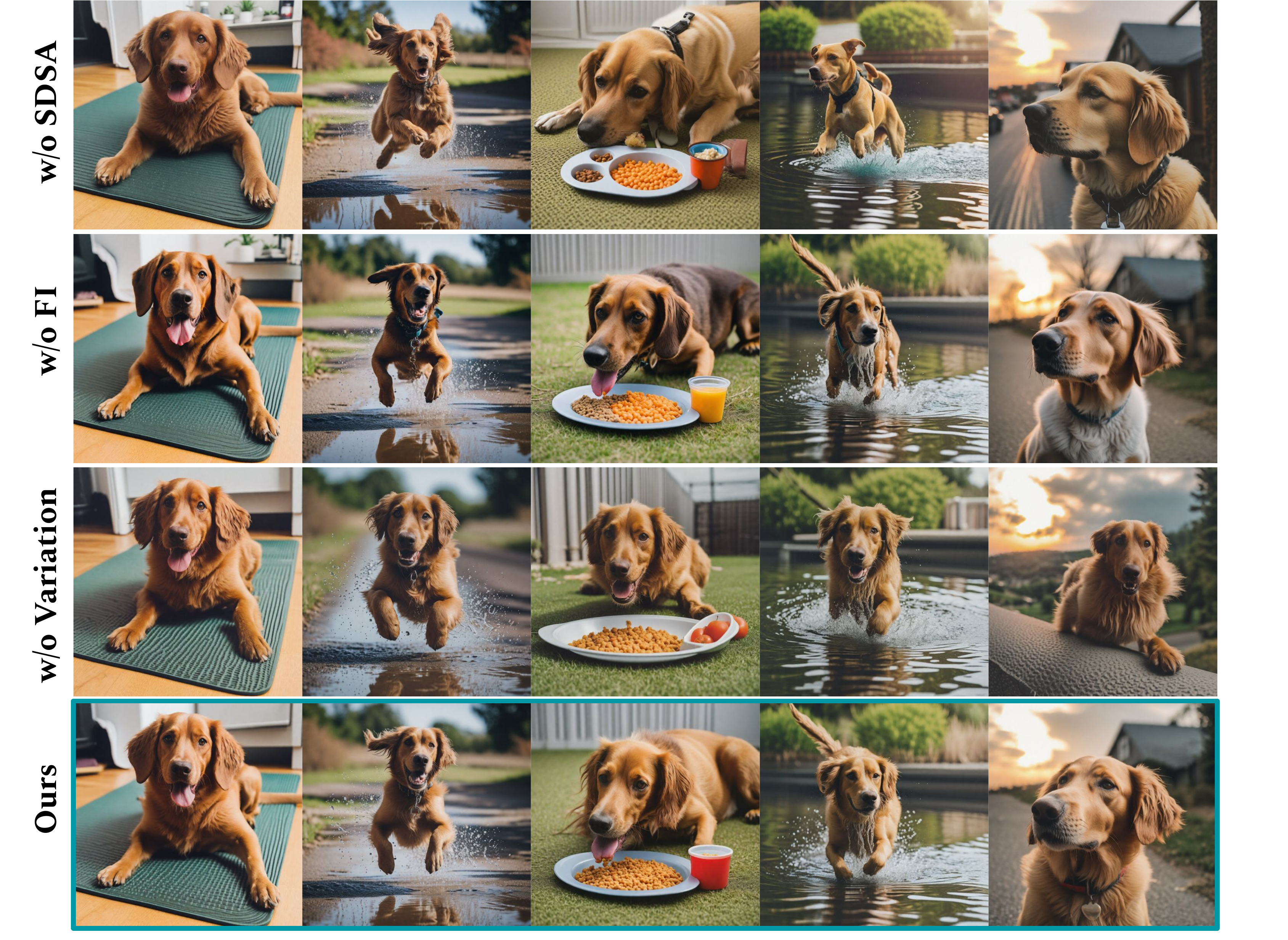} %
    \caption{ \textbf{Component Ablation} We ablated several components: Subject-Driven Self Attention (SDSA), Feature Injection (FI), and our variation enriching strategies: self-attention dropout and Query-feature blending (Variation). All ablated cases fail to maintain consistency as our method.}
    \label{fig:ablation_qual}
\end{figure}

\begin{figure}[!b]
    \centering
    \includegraphics[width=\columnwidth, trim={1cm 85cm 95cm 0cm},clip]{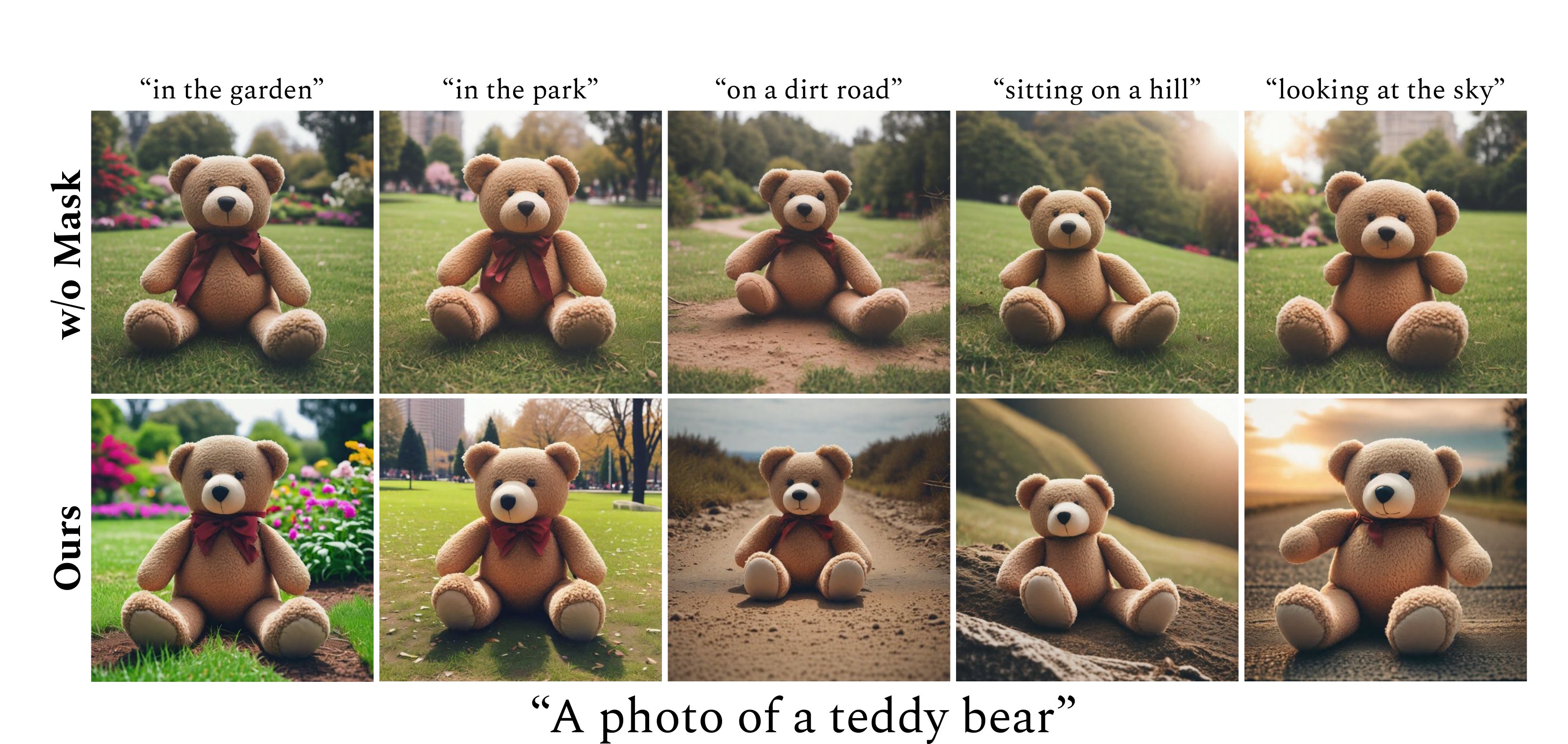} %
    \caption{ \textbf{Subject Masking:} Without the subject mask, there is noticeable background leakage across images.}
    \label{fig:mask_ablation}
\end{figure}

We now move to evaluate the impact of different components of our own method through an ablation study of the following components: \textbf{(1)} SDSA steps \textbf{(2)} Feature Injection (FI) \textbf{(3)} Attention dropout and query-feature blending. \textbf{(4)} Without using a subject mask. Qualitative comparisons are provided in Figures \ref{fig:ablation_qual}, \ref{fig:mask_ablation}. Quantitative results are provided in Appendix \ref{fig:ablation_appendix}. 

Removing SDSA leads to poor consistency, both regarding shape and texture. Removing FI yields subjects with similar shapes but less accurate identities. Finally, removing the vanilla query blending and attention dropout leads to significant reductions in layout diversity. We quantify this loss of diversity in Appendix \ref{sec:diversity}.

Additional examples on the effects of the feature injection are provided in \cref{fig_noisy_correspondence}. In cases where the subject can take forms with significantly different semantics (e.g., a humanoid robot versus a factory's robotic arm), the DIFT features may lead to noisy correspondences. While such noisy correspondence maps can lead to sub-optimal subject consistency, we observe that in the majority of cases, feature injection still contributes to more consistent results. We also highlight the method's ability to handle occluded objects (e.g. the robots legs, which do not appear in one of the frames). This is grounded in the thresholding mechanism, which drops patches with low similarity scores.

\subsection{Extended Applications}
\begin{figure}[!htb]
    \centering
    \includegraphics[width=0.9\columnwidth, trim={0.2cm 5cm 0cm 0cm},clip]{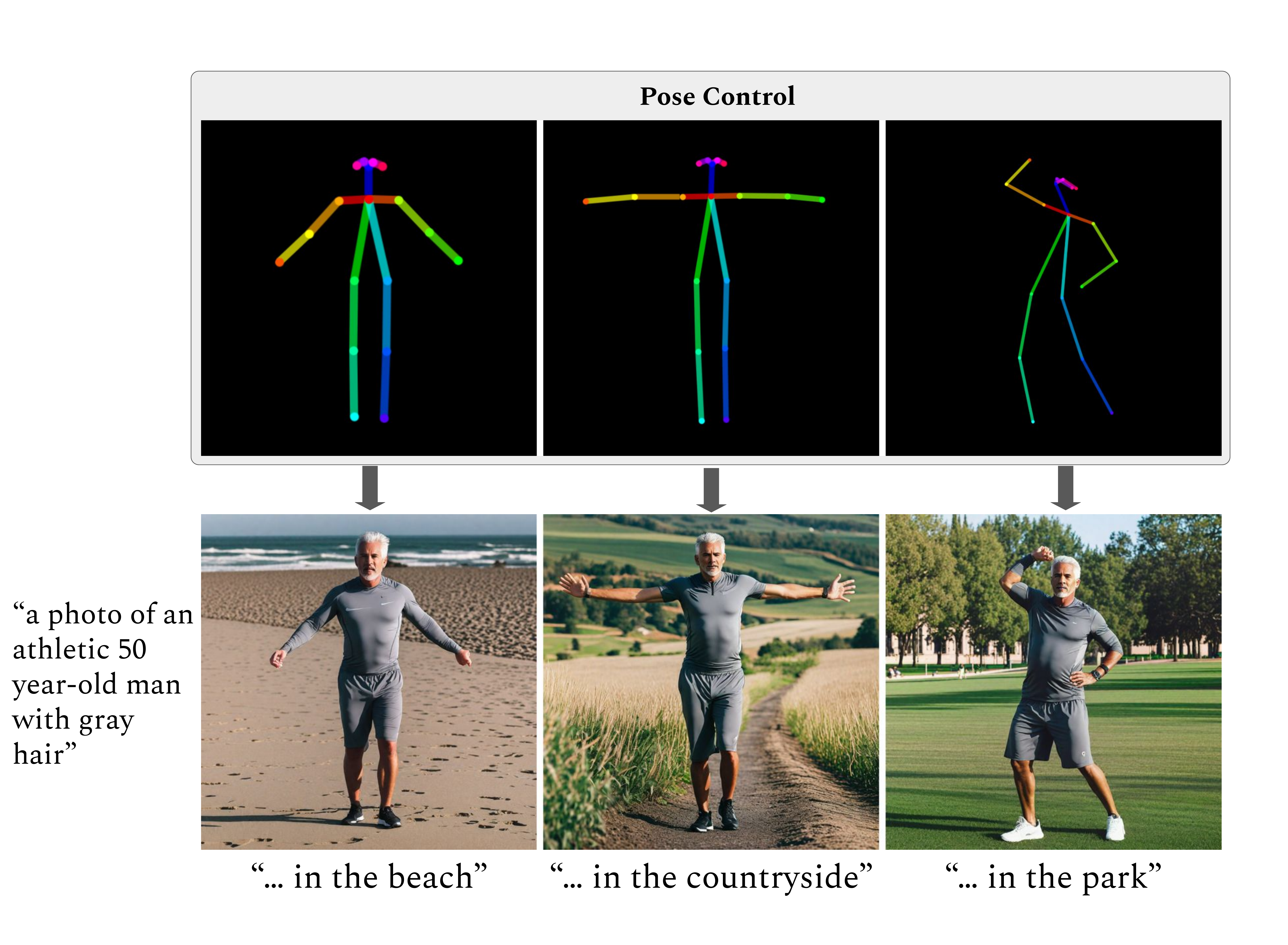} %
    \caption { \textbf{ControlNet Integration.} Our method can be integrated with ControlNet to generate a consistent character with pose control.}
    \label{fig:controlnet}
\end{figure}

\paragraph{Spatial Controls.}
We first demonstrate that \ourmethod is compatible with existing guided generation tools like ControlNet~\cite{zhang2023adding}. As these are compatible with standard personalization methods~\cite{gal2022textual,zhang2023adding,avrahami2023bas}, we want to ensure that our alternative approach maintains this compatibility.
In \cref{fig:controlnet} we show that pose-based controls with ControlNet successfully guide our consistent image generation method. 

\begin{figure}[!htb]
    \centering
    \includegraphics[width=0.9\columnwidth, trim={0.5cm 0cm 0.5cm 0cm},clip]{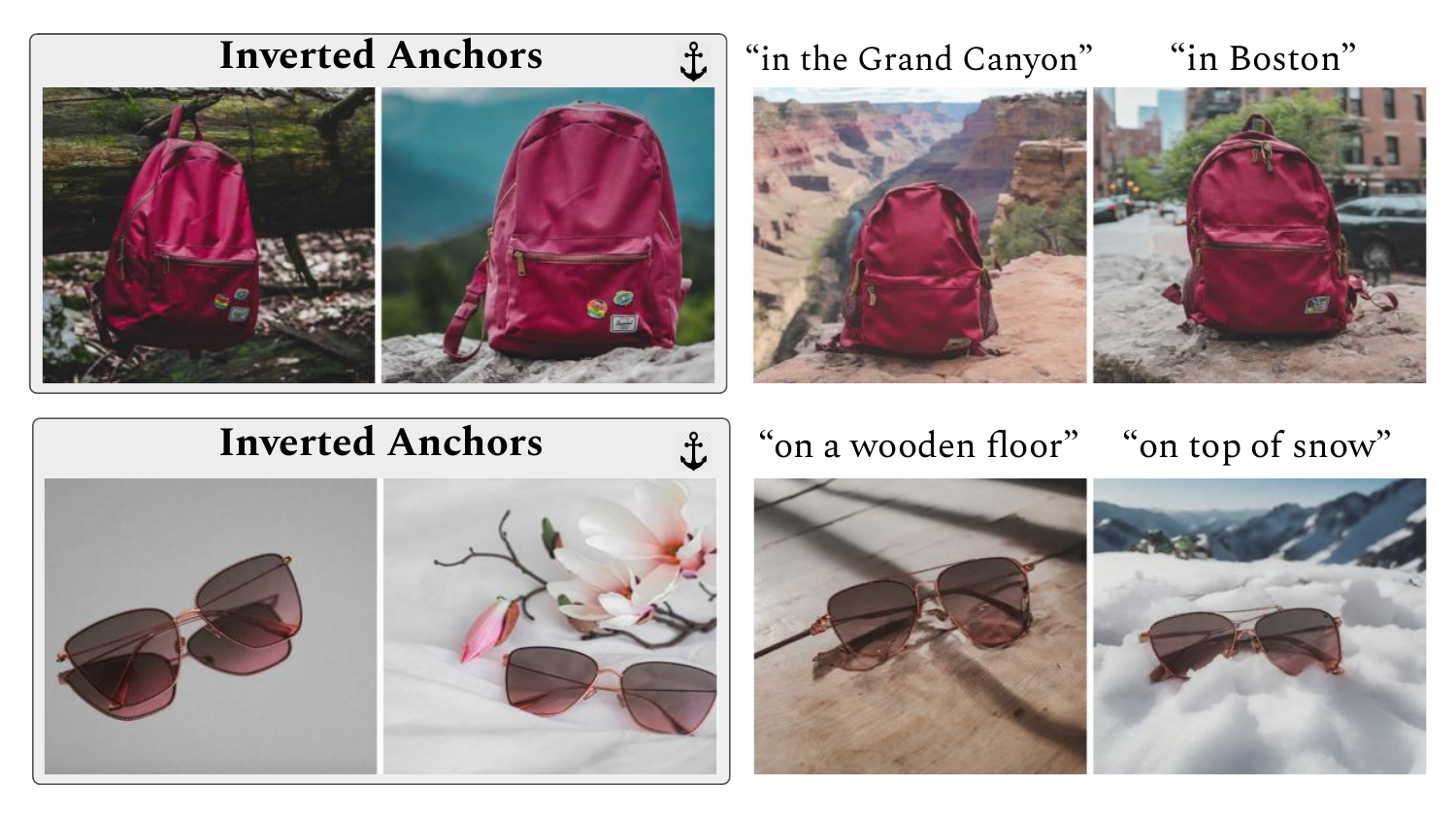} %
    \caption{ \textbf{Training-Free Personalization.}     We utilize edit-friendly inversion to invert 2 real images per subject. These inverted images used as anchors in our method for training-free personalization. }
    \label{fig:personalization}
\end{figure}

\paragraph{Training-free Personalization.}
We are \textit{the first to show} training-free \textit{personalization} (\cref{fig:personalization}), where \ourmethod enables personalization without any tuning or encoder use. Specifically, we show how to personalize common subject classes. Given two images of a subject, we invert them using Edit Friendly DDPM-Inversion~\cite{huberman2023edit}. These latents and noise maps are used as anchors for \ourmethod, allowing the rest of the batch to draw on their visual appearance. This application requires minor modifications of \ourmethod, which we detail in Appendix \ref{training_free_appendix}.

In \cref{fig:personalization_failure} we show that this approach struggles with complex objects, and is incompatible with style-changing prompts. Yet, we believe it can serve as a quick and cheaper alternative to current personalization methods, and leave it for future work.

\begin{figure}[!htb]
    \centering
    \includegraphics[width=0.85\columnwidth, trim={8cm 66cm 110cm 4cm},clip]{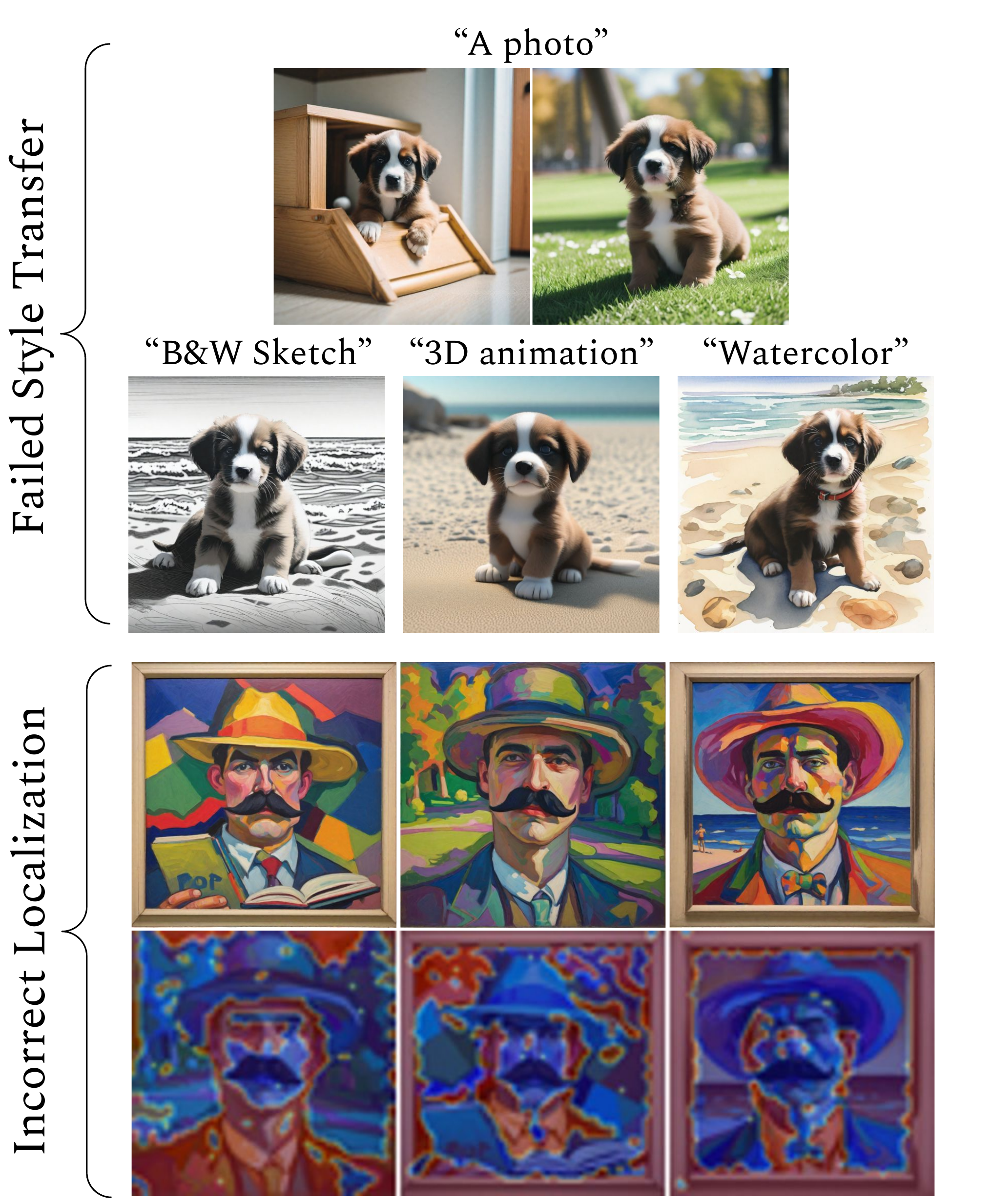} %
    \caption{ \textbf{Limitations:} Our method often struggles with different styles in the same set of image (Top), and is dependent on the quality of the model cross-attention to localize the subject correctly (Bottom).}
    \label{fig:limitations}
\end{figure}

\section{Limitations}

Our approach has several limitations, shown in \cref{fig:limitations}. First, it relies on the localization of objects through cross-attention maps. This process may occasionally fail, particularly when dealing with unusual styles. However, from our observations so far, such failures appear relatively infrequent (fewer than $5\%$), and they can be resolved by simply changing the seed. 
Another limitation is found in the entanglement between appearance and style. Our method struggles to separate the two, and hence we are limited to consistent generations where images share the same style.

\begin{figure}[!htb]
    \centering
    \includegraphics[width=1.\columnwidth, trim={10cm 83cm 80cm 2cm},clip]{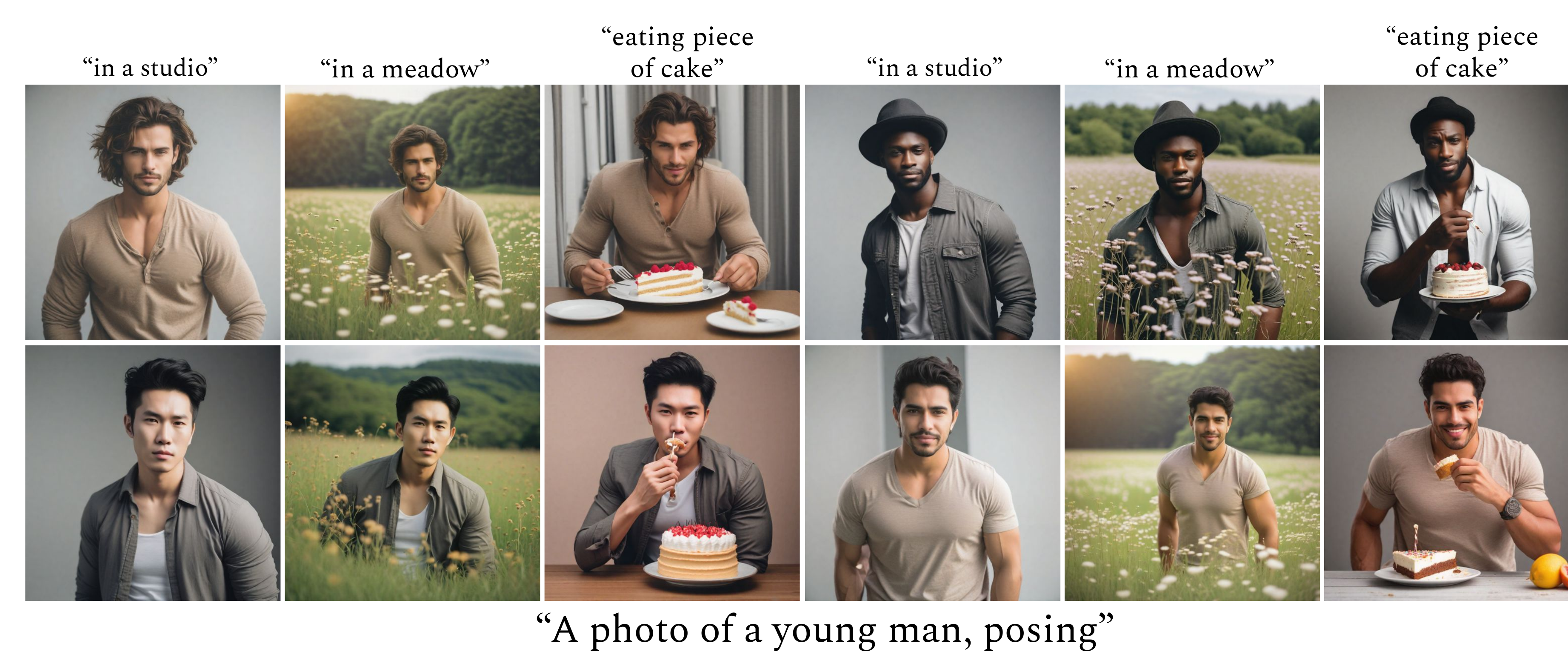} %
    \caption{\textbf{Model Bias.} The underlying SDXL model may exhibit biases towards certain ethnic groups, and our approach inherits them. Our method can generate consistent subjects belonging to diverse groups when these are highlighted in the prompt.}
    \label{fig:model_bias}
\end{figure}

Additionally, we observe that the underlying SDXL model may exhibit biases towards certain groups. We demonstrate that these biases can be significantly reduced by specifying modifiers like gender or ethnicity, as illustrated in \figref{fig:model_bias}.

\section{Conclusions}

We introduced \ourmethod, a training-free approach for creating visually consistent subjects using a pre-trained text-to-image diffusion model. When compared to the state of the art, our method is not only $\times 20$ faster, but can better preserve the output's alignment with the given prompts. Moreover, our method can be easily extended to tackle more challenging cases such as multi-subject scenarios, and even enable training-free personalization for common objects.

We hope that our results will assist in a consistent generation of creative endeavors and that they will inspire others to continue exploring training-free alternatives to personalization-based tasks.

\begin{acks}
    We thank Assaf Shocher, Eli Meirom, Chen Tessler, Dvir Samuel and Yael Vinker for useful discussions and for providing feedback on an earlier version of this manuscript. This work was completed as part of the first author's PhD thesis at Tel-Aviv University.
\end{acks}

\clearpage

\bibliographystyle{ACM-Reference-Format}
\bibliography{main}

\clearpage

\appendix

\renewcommand{\thefigure}{A.\arabic{figure}}
\setcounter{figure}{0}
\textbf{\huge Appendix: Training-Free Consistent Text-to-Image Generation}
\section{Additional results}
\begin{figure*}[!ht]
    \centering
    \includegraphics[width=0.8\textwidth, trim={10.0cm 30.0cm 52.0cm 0.cm},clip]{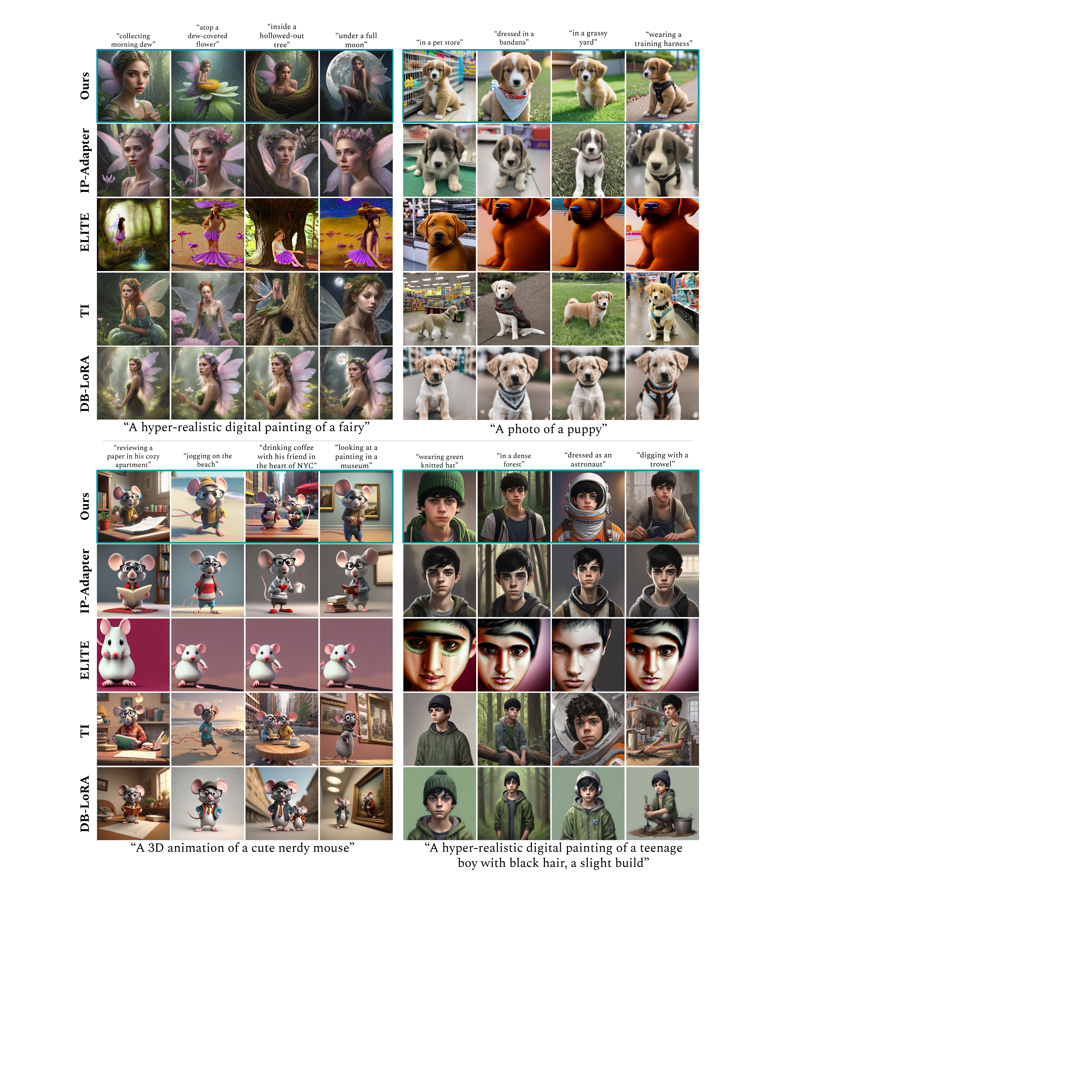} %
    \caption{
    \textbf{Additional Qualitative Comparisons}  We evaluated our method against IP-Adapter, TI, ELITE and DB-LORA.
    Some methods failed to maintain consistency (TI) or follow the prompt (IP-Adapter). Other methods alternated between keeping consistency or following text, but not both (DB-LoRA). Our method successfully followed the prompt while maintaining consistency.}
    
    \label{fig:fig_extra_qualitative_baselines}
\end{figure*}

\begin{figure*}[!ht]
    \centering
    \includegraphics[width=0.99\textwidth, trim={5.0cm 57.0cm 57.0cm 0.cm},clip]{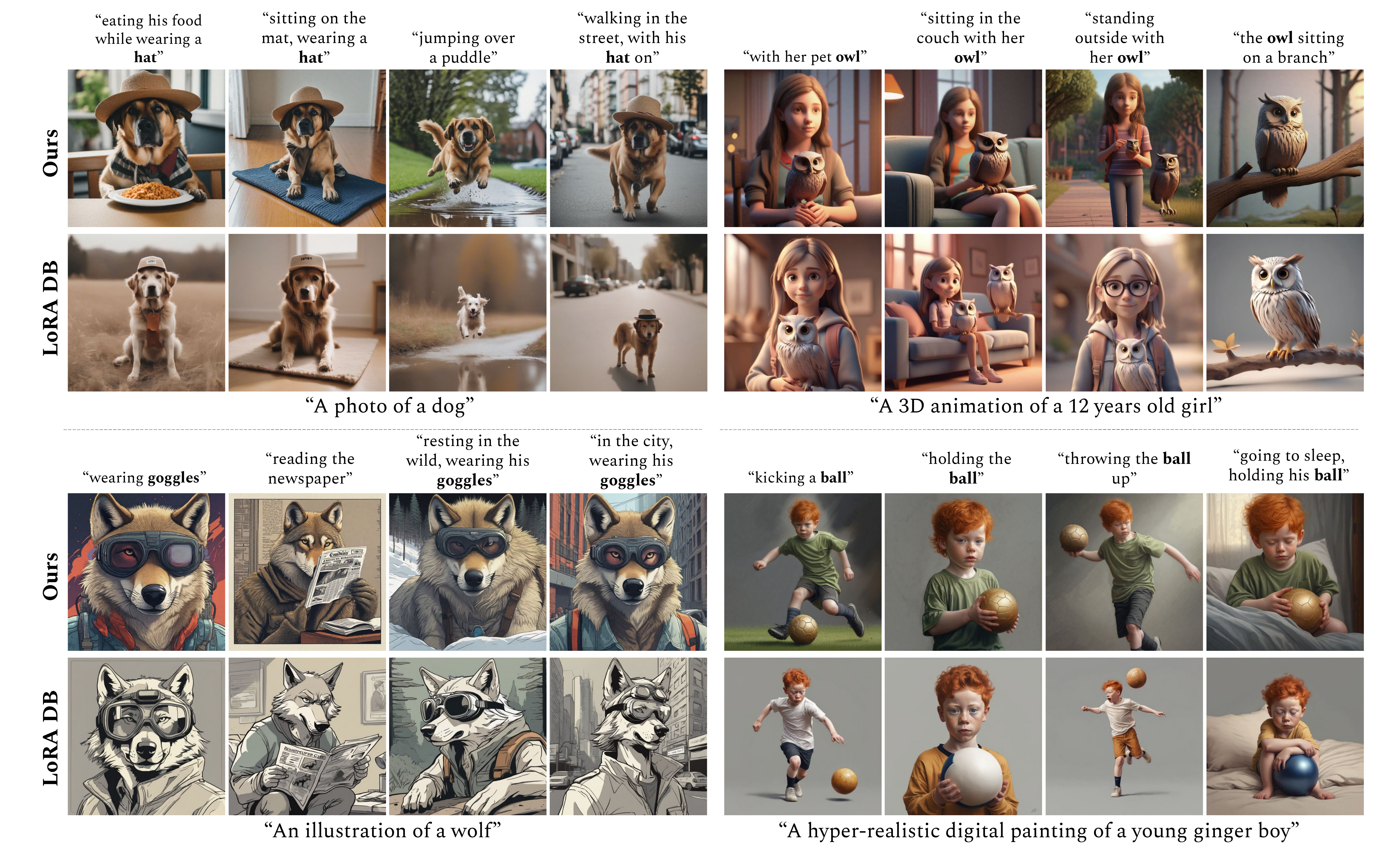} %
    \caption{
    \textbf{Additional Qualitative Multi-Subject Comparisons}  We evaluated our method against DB-LORA for generation of multiple consistent subjects. Lora DB tends to neglect the consistency of at least one of the subjects, while our method succeeds in both.
    }
    \label{fig:fig_extra_qualitative_vs_dblora}
\end{figure*}

\begin{figure}[htb]
    \centering
    \includegraphics[width=\columnwidth, trim={1cm 86cm 108cm 0cm},clip]{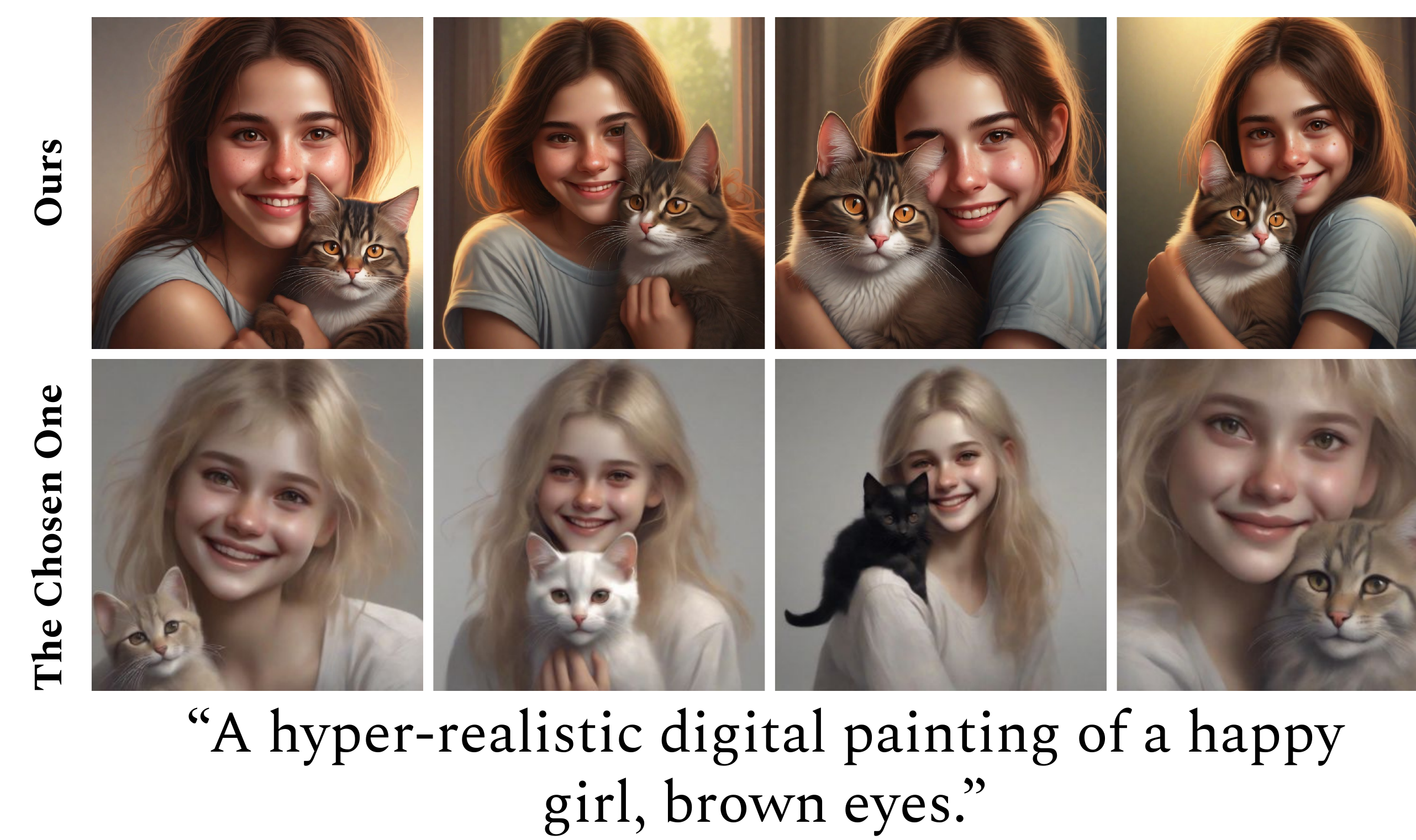} %
    \caption{ \textbf{Comparison to The Chosen One, Multi-Subject} We evaluated our method against a concurrent work by \cite{avrahami2023chosen}. Notably, while \ourmethod is training-free, in contrast to the concurrent work that requires an iterative optimization process, we demonstrate that our method outperforms it in preserving consistency for multiple subjects.}
    \label{fig:chosen_one_multi}
\end{figure}

\begin{figure}[htb]
    \centering
    \includegraphics[width=\columnwidth, trim={0cm 0cm 0cm 0cm},clip]{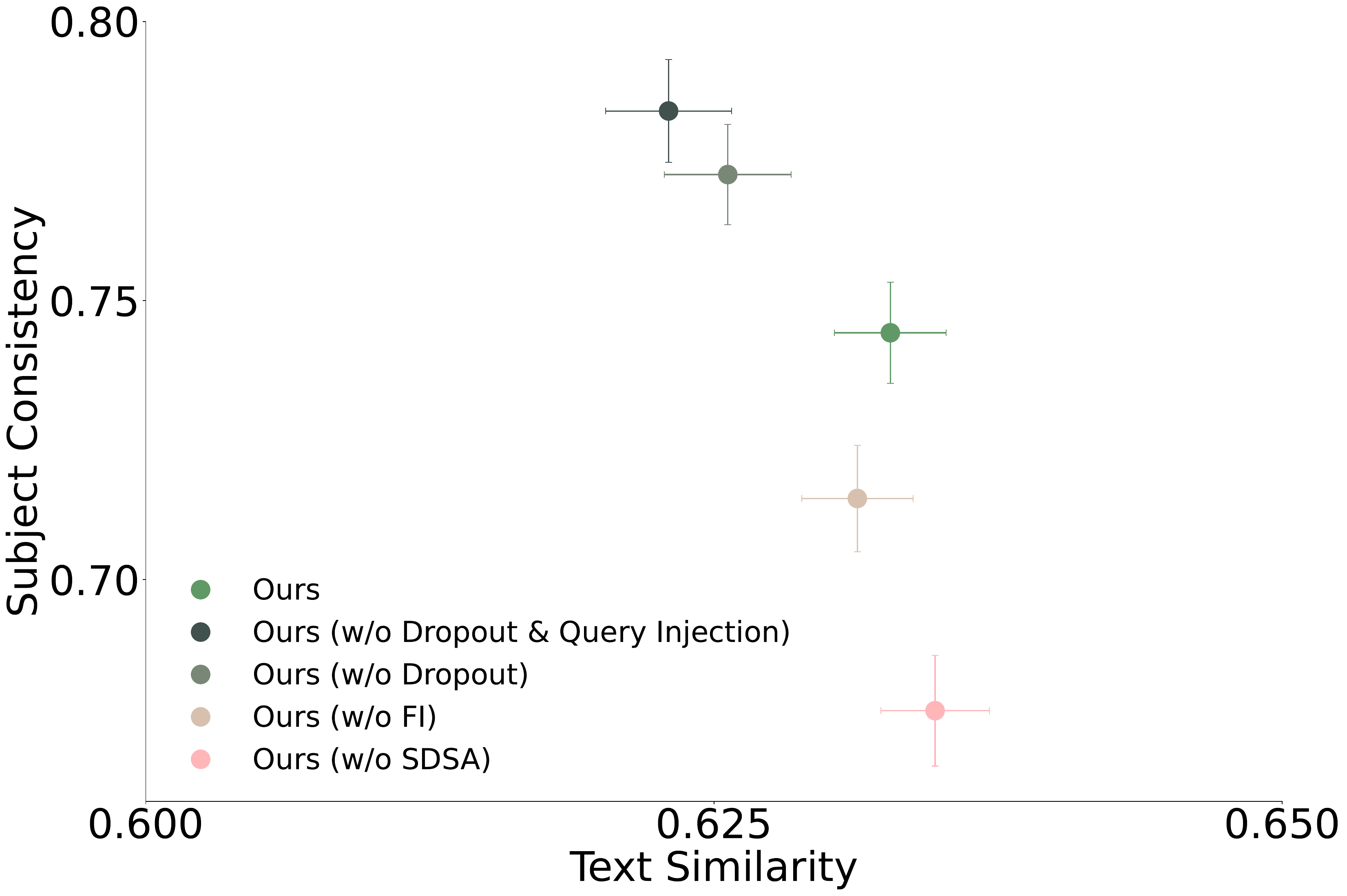} %
    \caption{ \textbf{Quantitative Component Ablation} We conducted a quantitative evaluation of \ourmethod by ablating various components: Self-Attention Dropout (Dropout), Query Injection, Feature Injection (FI), and Subject-Driven Self-Attention (SDSA). Notably, omitting either SDSA or FI resulted in reduced subject consistency. Eliminating the variation enhancement mechanisms (Dropout and Query Injection) decreased textual similarity.}%
    \label{fig:ablation_appendix}
\end{figure}

\begin{figure}[htb]
    \centering
    \includegraphics[width=\columnwidth, trim={0cm 0cm 0cm 0cm},clip]{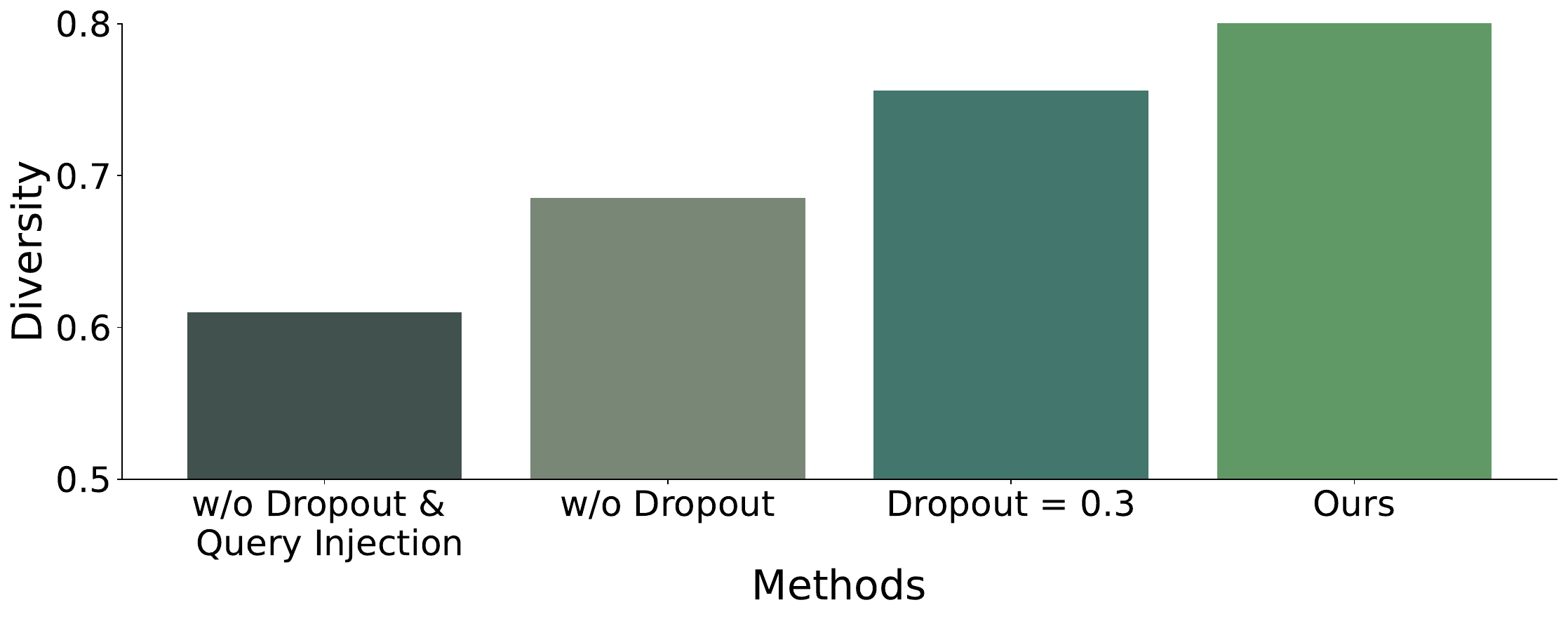} %
    \caption{ \textbf{Layout Diversity Ablation} Our method attains the highest Diversity Score, when compared to variations omitting the Dropout and/or the Query Injection components.}
    \label{fig:diversity}
\end{figure}

We provide additional Qualitative and Quantitative results of our method. In \cref{fig:fig_extra_qualitative_baselines}, \cref{fig:fig_extra_qualitative_vs_dblora} and \cref{fig:chosen_one_multi} we show additional single and multi subject qualitative comparisons to existing baselines. In \cref{fig:ablation_appendix}
 and \cref{fig:diversity} we provide quantitative ablation results with respect to Textual Similarity, Subject Consistency, and Layout Diversity.

\section{Additional implementation details}
\label{supp_implementation_details}

\paragraph{Subject-Driven Self-Attention}
is applied at all timesteps, in the decoder layers of the U-net.

\paragraph{Feature injection and DIFT features:}
Feature injection is applied at timesteps $t \in [680, 900]$, with $\alpha = 0.8$. We only inject patches with similarity scores above a threshold that is set automatically by the Otsu method.

 For feature injection, we denoise the image from $t=1000$ to $t=261$, for computing the DIFT features at $t=261$ as in \cite{tang2023emergent}. We then denoise again from $t=1000$ to $t=0$, guiding the feature injection with the precomputed DIFT features.

\paragraph{Pose Variation}
We apply the injection of Vanilla Query Features over the first $5$ denoising steps, with $\nu_t$ values that linearly decay with $t$ from $0.9$ to $0.8$.  Self-Attention Dropout is applied with $p=0.5$.

\paragraph{Diffusion Process}
Images were sampled with $50$ DDIM steps and a guidance scale of $5$.
Similarly to \cite{alaluf2023crossimage, luo2023knowledge}, we used Free-U \cite{si2023freeu} to enhance the generation quality.

\paragraph{Extracting per subject mask}
To extract subject masks from noisy latent patches,  we collect all cross-attention maps that relate to each subject's token, across all previous diffusion steps, and all cross-attention layers of resolution $32\times32$. We then average them, and threshold them using the ``Otsu's method'' \cite{otsu}. The subject mask at the generation step $\tau$, is given by:
\begin{align}
m_i &= \mathbb{E}_{l=0}^{L} \mathbb{E}_{t=T}^{\tau} \left[ A_i \right] \nonumber \\
    M_{i} &= otsu\Big( m_{i} \Big) \in \{0,1\}^{P} ,
\end{align}
where $L$ is the number of network layers, $P$ is the number of patches and $\mathbb{E}$ denotes averaging.

\paragraph{Constructing DIFT pairwise correspondence maps}
For a source image $I_s$ and a target image $I_t$, we denote their DIFT features by $D_s$, $D_t$ $\in \R^{P \times d_{DIFT}}$ respectively, where $d_{DIFT}$ is the feature dimension. The cross-image patch similarity scores are given by the cosine similarity between these features:
\begin{equation}
    \text{Sim}(I_s, I_t)_{p_s, p_t} = \frac{D_s[p_s] \cdot D_t[p_t]}{\lVert D_s[p_s] \rVert~\lVert D_t[p_t] \rVert},
\end{equation}
where $p_s$, $p_t$ are the indices of specific patches in the source and target image respectively and $D_s[p_s]$, $D_t[p_t]$ are the DIFT matrix rows (\ie feature-vectors) matching these patches. Given these patch-similarity scores, we can calculate a patch-wise dense-correspondence map $C_{t \rightarrow s}$: 
\begin{equation}
\forall p_t \in I_t:\,\,~~C_{t \rightarrow s}[p_t] = \argmax_{p_s \in I_s} \text{Sim}(I_s, I_t)_{p_s,p_t}.
\end{equation}
Here, for each patch $p_t$ in the target image $I_t$, $C_{t \rightarrow s}[p_t]$ gives the index of the most similar patch in the source image, $I_s$. 

\paragraph{Control Net} To enhance the impact of the pose from ControlNet, we raise the Self-Attention Dropout value to 0.7.

\section{Diversity Evaluations} \label{sec:diversity}
In our quest to enhance layout diversity, we have proposed two strategies: Vanilla Query Injection and Self-Attention Dropout. To quantitatively assess the contributions of these strategies, we construct an automatic evaluation metric for layout diversity.

Specifically, we utilize the DIFT features and dense correspondence maps between image pairs. We measure layout diversity by calculating the average displacement between corresponding points across each image pair. A low displacement score indicates that the subject's layout is largely aligned in both images. If this persists across the entire image set, we can conclude that the images have limited diversity. Notably, we opt for a geometric-diversity metric rather than an image-based one since the latter may conflate layout diversity with undesired changes in the subject's appearance. In \cref{fig:diversity}, we show the results of our diversity metric when ablating components from our method. Scores are normalized to the displacement value of images sampled from a vanilla non-consistent SDXL model. Our full method attains high diversity scores, surpassing solutions that omit one or more of the layout enrichment components.

\section{Prompts Dataset Details} \label{sec_supp_additional_details}
To evaluate our method on a large scale, we constructed a prompts dataset. We asked ChatGPT to construct sets of 5 prompts each, where every prompt in a set contains the same recurring subject. In addition, we instructed it to generate the following metadata for each prompt set:

\begin{itemize}
    \item \textbf{Superclass:} We asked the model to group each set into one of the following superclasses: humans, animals, fantasy, inanimate.
    \item \textbf{Subject Token:} A single token representing the subject (e.g., 'cat', 'boy').
    \item \textbf{Subject Description:} The description of the recurring subject (e.g., "A 16-year-old girl with blonde hair").
    \item \textbf{Description Level:} This refers to whether the subject description is generic or detailed. For example, a generic description might simply be "A dog", while a detailed description could elaborate as "A dog with brown and white fur, fluffy hair, and pointy ears".
    \item \textbf{Style:} The style of the requested image; for example, "A 3D animation of."
\end{itemize} 
We generated 100 prompt sets, comprising a total of 500 images, spanning various superclasses, description levels, and styles. We aimed to cover a wide spectrum of visual themes and subjects, ensuring a broad and inclusive representation across the different categories.

Our prompt dataset is provided as a YAML file in the supplemental materials.

\section{Training-Free Personalization Details} \label{training_free_appendix}

We implemented edit-friendly inversion \cite{huberman2023edit} for SDXL and set the guidance scale to 2.0. We inverted two real-images, as shown at \cref{fig:personalization}, and used them as anchors in our method. However, in our method, the anchors are generated with attention sharing between them, which is not suitable for the case of inverted anchors. Therefore, we modify our anchoring process such that anchors will only share attention maps with generated images, rather than with other anchors. We observed that naively using the inverted image attention features with our method gave poor personalization results, and we hypothesize it is related to a distribution shift between the features of the inverted and generated images. Hence, we added Adain \cite{adain_huang2017arbitrary} to align the self-attention keys of inverted features with the self-attention keys of the generated images. 

\begin{figure}[htb]
    \centering
    \includegraphics[width=0.9\columnwidth, trim={0cm 35cm 0cm 0cm},clip]{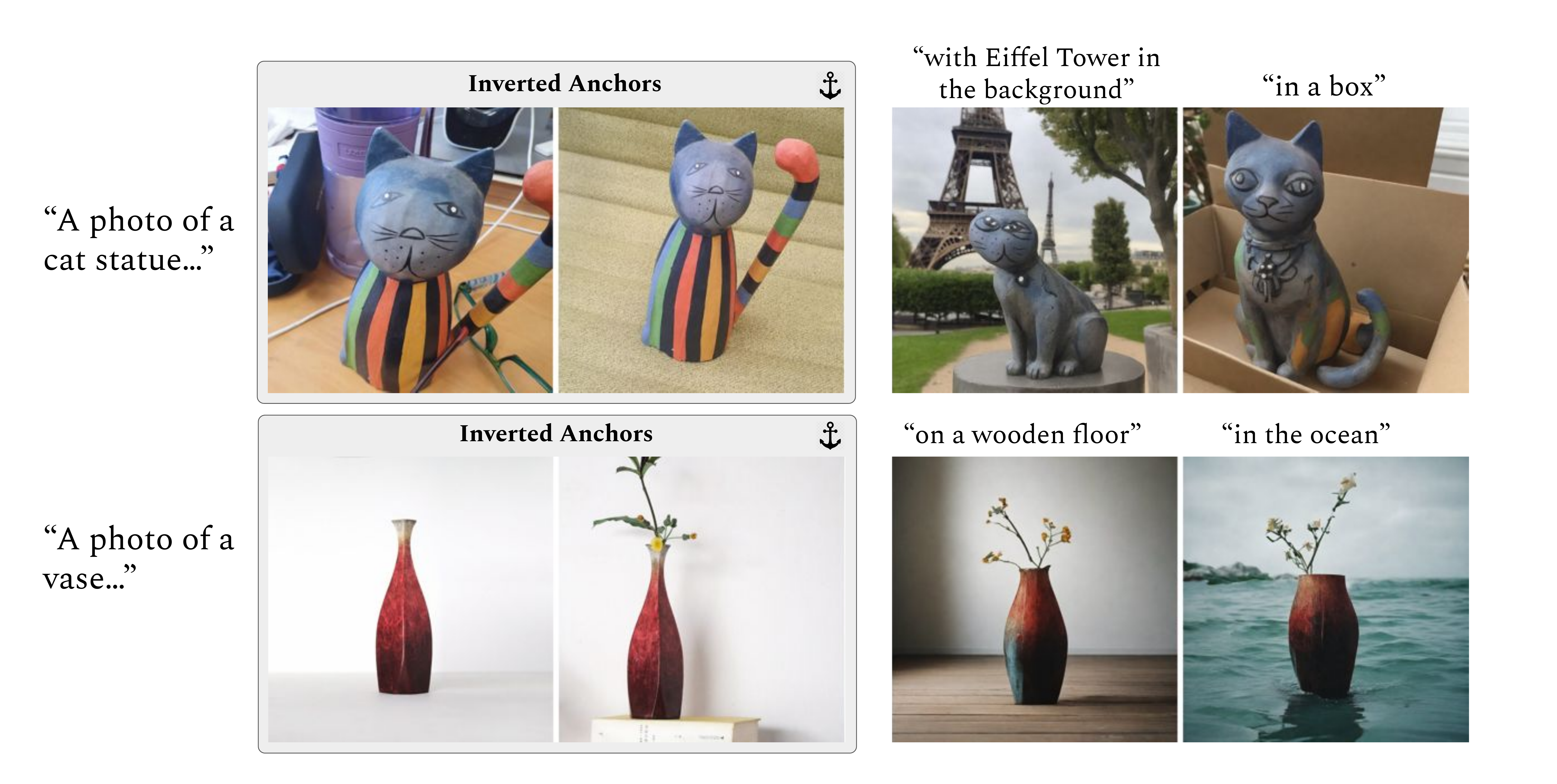} %
    \caption{ \textbf{Training-Free Personalization Failure} Our training-free personalization method may fail on non common subjects. }
    \label{fig:personalization_failure}
\end{figure}

Since edit-friendly inversion is based on DDPM, we modified our generation process to also use DDPM scheduling with 100 generation steps. We used the default guidance scale of 5.0 for the generated images. We note that this training-free personalization only works for simple objects, and provide failure cases at \cref{fig:personalization_failure}. 

\section{User Study Details}
\label{suppl_user_study}
We evaluate the models through two Amazon Mechanical Turk user studies using a two-alternative forced choice protocol. In the first study, named \textbf{``Visual consistency"}, raters saw two image sets, each produced by a different approach. They chose the set where the subject's identity remained most consistent. In the second study named \textbf{``Textual alignment"}, users received a text description and two images from different approaches. They selected the image that better matched the description.

\subsection{Visual Consistency}
\begin{figure*}[ht]
\includegraphics[width=0.9\textwidth, trim={0.cm 0.cm 0.cm 0.cm},clip]{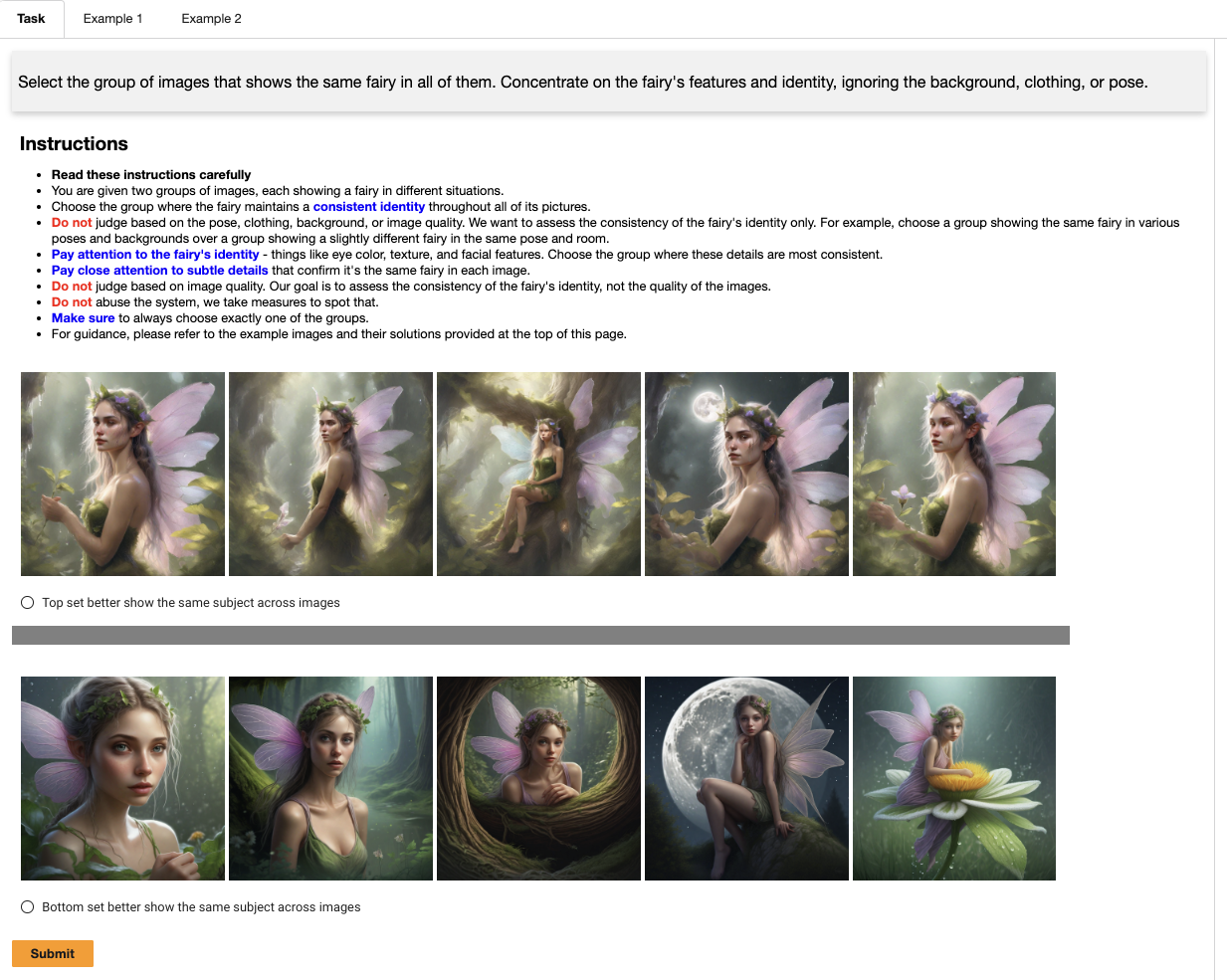} %
\caption{One trial of the visual consistency user study.}
\label{fig_amt1_task}

\begin{flushleft}
\includegraphics[width=0.48\textwidth, trim={0.cm 0.cm 0.cm 0.cm},clip]{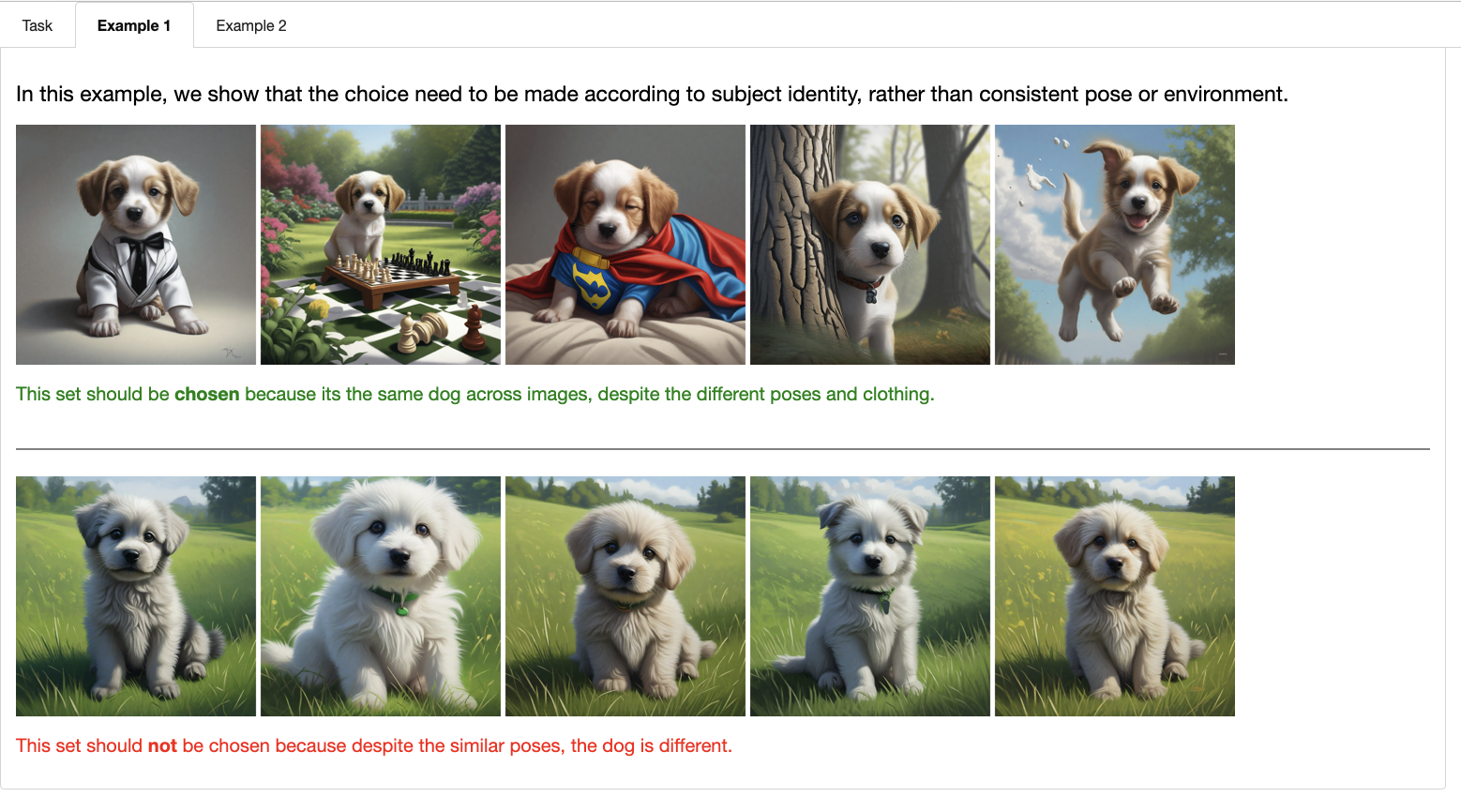} %
\includegraphics[width=0.48\textwidth, trim={0.cm 0.cm 0.cm 0.cm},clip]{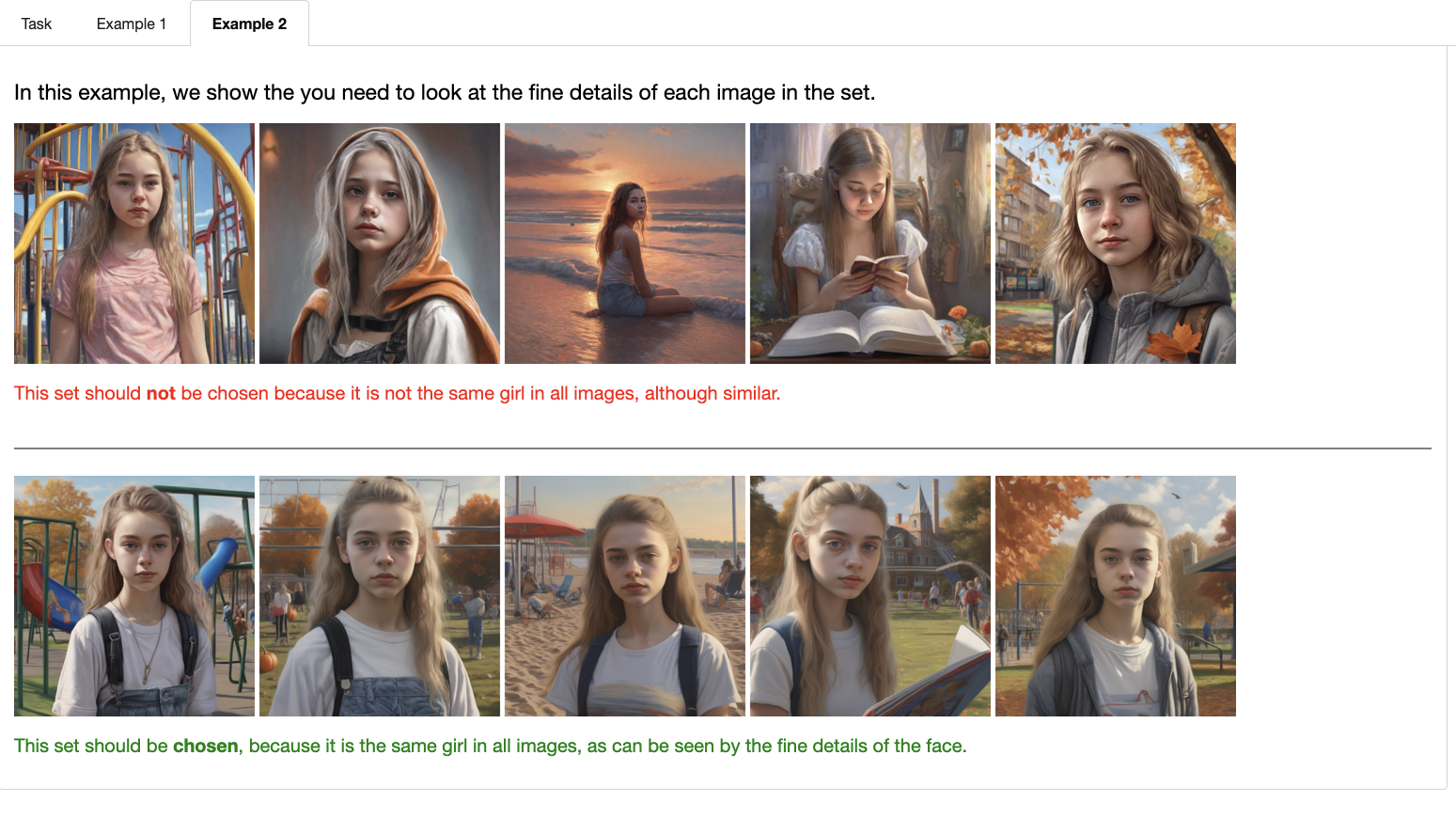} %
\caption{Examples provided in the visual consistency user study. }
\label{fig_amt1_examples}
\end{flushleft}
\end{figure*}

For the first study, in each trial, raters were presented with two sets of images, each set depicting a subject in various situations. They were instructed to select the set where the subject's identity remained consistent across all images. This decision was to be based solely on the subject's features and identity, disregarding elements such as background, clothing, or pose. The focus was on the consistency of the subject's identity, including aspects like eye color, texture, facial features, and other subtle details. Example images and solutions were provided for guidance.
Figure \ref{fig_amt1_task} illustrates the experimental framework used in the trials. Figure \ref{fig_amt1_examples} displays the example images provided to help guide raters through the instructions.

The study included 100 trials of images from 100 unique subjects. Each trial was repeated 5 times with 5 different raters. 
Each image set consisted of 5 images generated from 5 distinct prompt settings. 
For the \ourmethod images, we selected the variant with less stringent identity preservation parameters.
Raters were paid $0.15$ per trial. To maintain the quality of the study, we only selected raters with an Amazon Mechanical Turk  ``Masters'' qualification, demonstrating a high degree of approval rate over a wide range of tasks. Furthermore, we also conducted a qualification test on the prescreened pool of raters, consisting of 5 curated trials that were simple.

\subsection{Textual Alignment}
\begin{figure*}[ht]
\includegraphics[width=\textwidth, trim={0.cm 0.cm 0.cm 0.cm},clip]{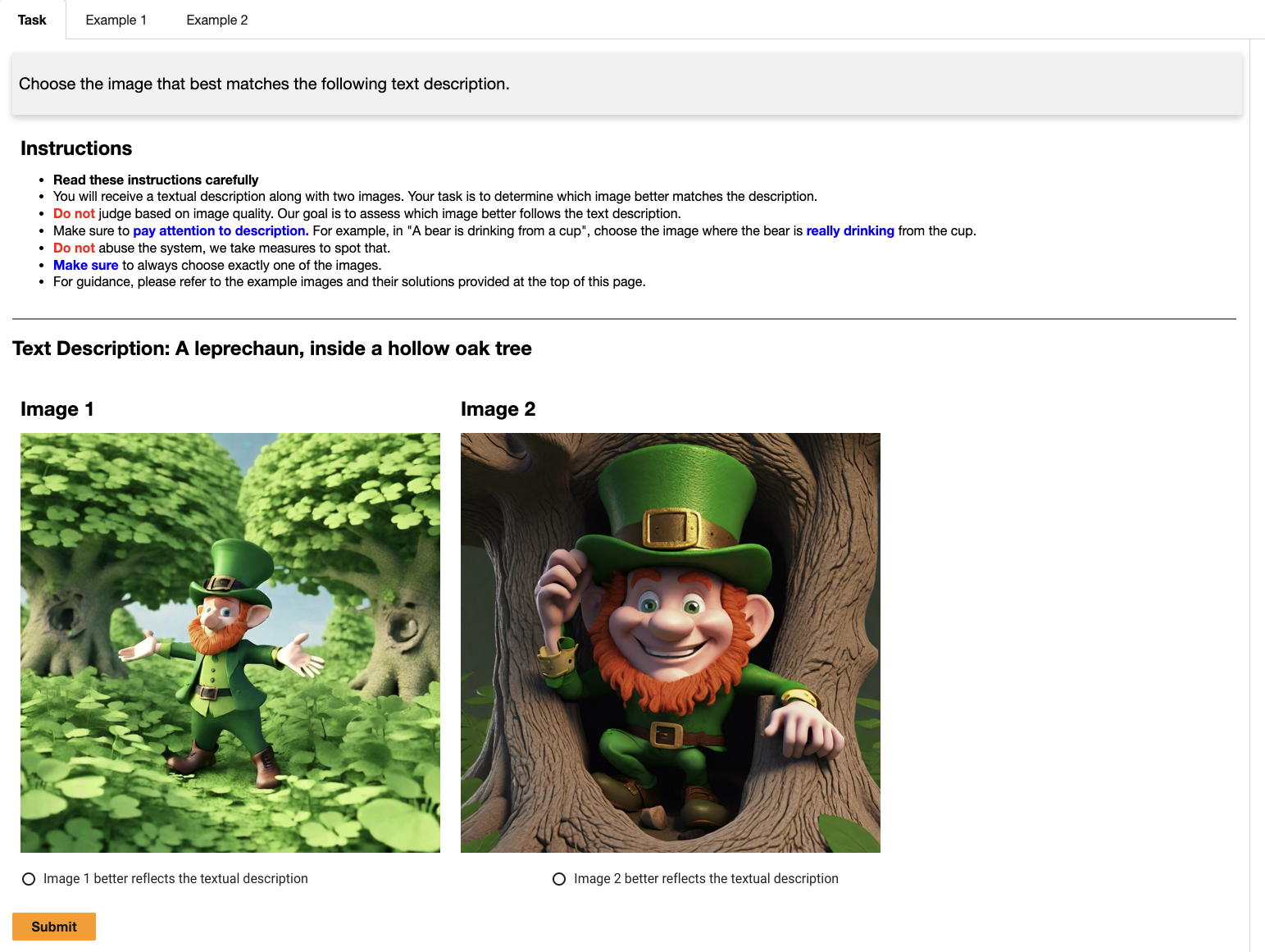} %
\caption{One trial of the textual alignment user study. }
\label{fig_amt2_task}

\begin{flushleft}
\includegraphics[width=0.48\textwidth, trim={0.cm 0.cm 0.cm 0.cm},clip]{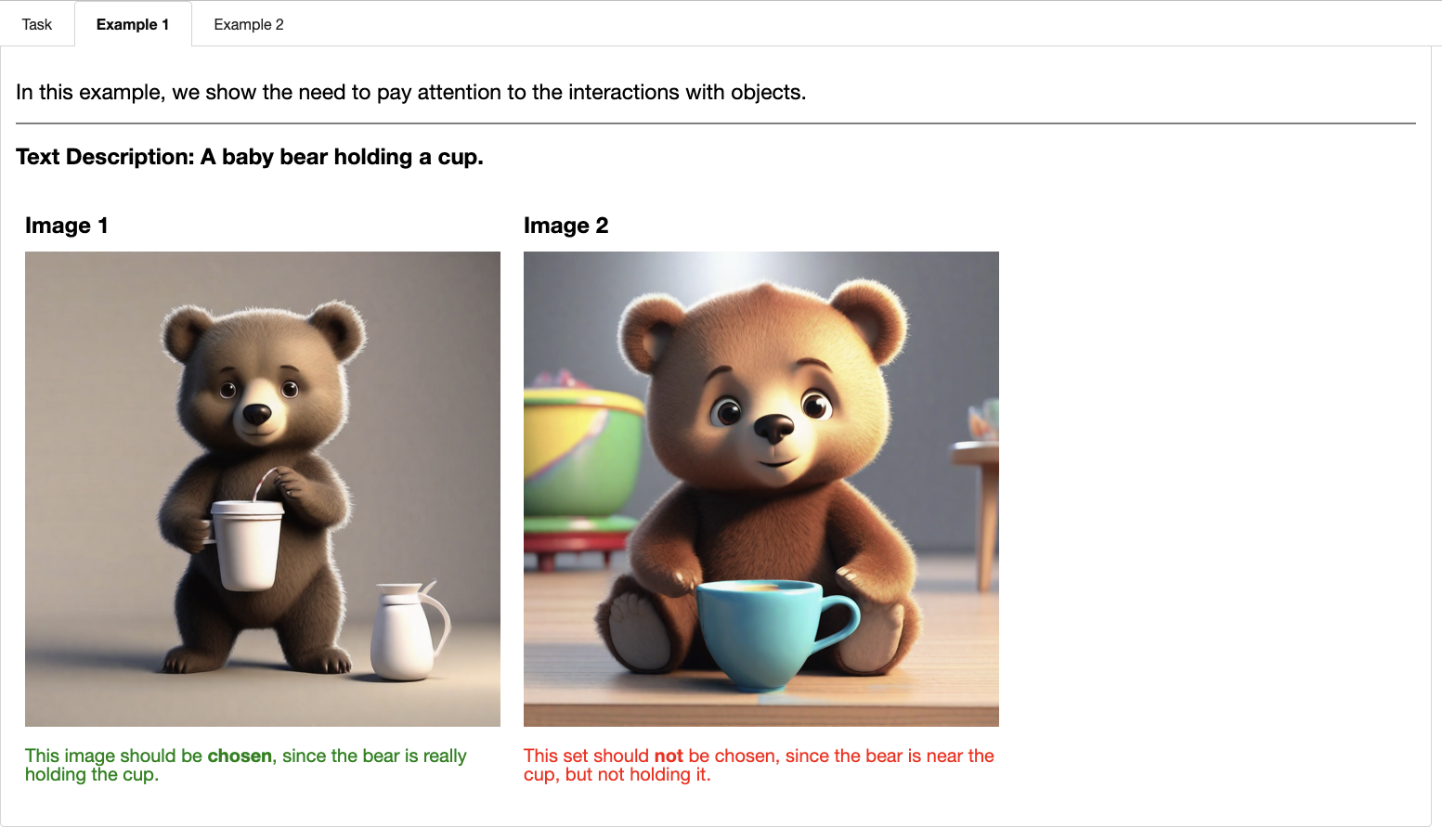} %
\includegraphics[width=0.48\textwidth, trim={0.cm 0.cm 0.cm 0.cm},clip]{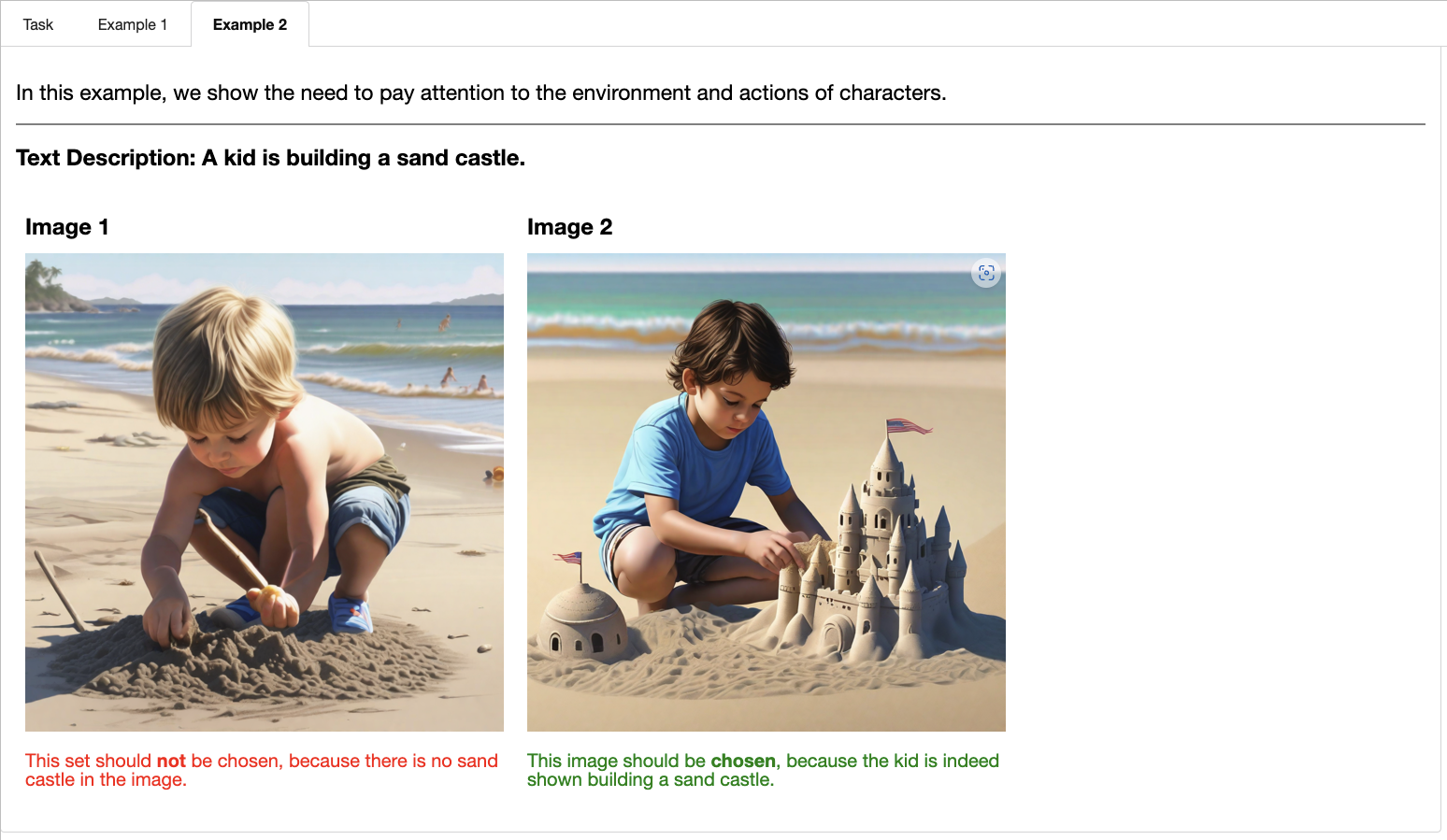} %
\caption{Examples provided in the textual alignment user study. } 
\label{fig_amt2_examples}
\end{flushleft}
\end{figure*}

For the second study, in each trial, raters were presented with a textual description and two images. They were instructed to determine which image better matched the textual description. They were advised to focus on the details of the description. For example, if the text states "A bear is drinking from a cup", raters should choose the image where the bear is actually drinking from the cup. For guidance, example images with solutions were provided.
\figref{fig_amt2_task} illustrates the experimental framework that was used in trials. The text descriptions only included the general subject class (\eg dog'') rather than fully detailed descriptions (\eg A cute brown and white fluffy puppy dog with blue eyes''). This helped raters focus on the subject's setting. \figref{fig_amt2_examples} displays the examples provided to help guide raters through the instructions.

The study included 500 trials of images from 100 unique subjects, each subject had 5 distinct prompt settings.  
For the \ourmethod images, we selected the variant with less stringent identity preservation parameters.
We paid \$0.05 per trial. To maintain the quality of the study, we only selected raters with an Amazon Mechanical Turk  ``Masters'' qualification. Furthermore, we also conducted a qualification test on the prescreened pool of raters, consisting of 5 curated trials that were simple.

\end{document}

%% file: macros.tex
\input{math_commands}

\usepackage{amsmath}
\usepackage{comment}
\usepackage{multirow,bigdelim}
\usepackage{lipsum}
\usepackage{array}
\usepackage{wrapfig}

\usepackage[percent]{overpic}
\usepackage{makecell}
\usepackage{blindtext}
\usepackage{xcolor}
\usepackage{soul}
\usepackage{cleveref}
\usepackage{amsfonts}
\usepackage{xspace} %
\usepackage{braket}
\usepackage{nicefrac}
\usepackage{enumitem} %
\usepackage{multicol} %
\usepackage{subfig} 
\usepackage{dsfont}

\makeatletter
\newcommand{\settitle}{\@maketitle}
\makeatother

\newcolumntype{C}[1]{>{\centering\let\newline\\\arraybackslash\hspace{0pt}}m{#1}}

\newif\ifdraft
\drafttrue

\newcommand{\tildeapprox}{{\raise.17ex\hbox{$\scriptstyle\sim$}}}

\renewcommand{\figref}[1]{\Figref{#1}}
\renewcommand{\cref}[1]{Fig. \ref{#1}}

\renewcommand{\eqref}[1]{Eq.~(\ref{#1})}

\definecolor{darkpink}{rgb}{0.561, 0.282, 0.427}
\definecolor{atomictangerine}{rgb}{0.8, 0.2, 0.1}
\definecolor{turq}{rgb}{0.0, 0.5, 0.5}
\definecolor{darkturq}{rgb}{0.0, 0.4, 0.4}
\definecolor{bright}{rgb}{0.8, 0.1, 0}
\definecolor{darkgray}{gray}{0.3}
\definecolor{gray}{gray}{0.5}
\definecolor{mahogany}{rgb}{0.6, 0.05, 0.05}
\definecolor{editblue}{rgb}{0.3,0.05,0.9}
\definecolor{black}{rgb}{0.,0.,0.}
\definecolor{darkgreen}{rgb}{0.1,0.5,0.0}
\definecolor{olive}{rgb}{0.537, 0.627, 0.318}
\definecolor{green}{rgb}{0.22, 0.463, 0.114}
\definecolor{grey}{rgb}{0.4, 0.4, 0.4}
\definecolor{blue}{rgb}{0.435, 0.659, 0.863}
\definecolor{pink}{rgb}{0.761, 0.482, 0.627}
\definecolor{darkpink}{rgb}{0.561, 0.282, 0.427}

\ifdraft

\newcommand\gal[1]{\textcolor{bright}{\textbf{GAL:} #1 }}
\newcommand\yuval[1]{\textcolor{pink}{\textbf{YUVAL:} #1 }}
\newcommand\yoad[1]{\textcolor{turq}{\textbf{YOAD:} #1 }}

\newcommand{\drop}[1]{}

\else
\newcommand{\dcc}[1]{}
\newcommand{\rgc}[1]{}
\newcommand{\opc}[1]{}
\newcommand{\gcc}[1]{}
\newcommand{\hmc}[1]{}
\newcommand{\abc}[1]{}

\newcommand\gal[1]{}
\newcommand\yuval[1]{}
\newcommand\yoad[1]{}
\fi

\newcommand{\ourmethod}{\textit{ConsiStory}\xspace}
\newcommand{\TTI}{T2I\xspace}

\def\Naive{Na\"{\i}ve\xspace}

\makeatletter
\DeclareRobustCommand\onedot{\futurelet\@let@token\@onedot}
\def\@onedot{\ifx\@let@token.\else.\null\fi\xspace}

\def\eg{\emph{e.g}\onedot}

\def\ie{\emph{i.e}\onedot}

\makeatother

\usepackage[bottom]{footmisc}
\usepackage{arydshln}

\makeatletter
\def\blfootnote{\xdef\@thefnmark{}\@footnotetext}
\makeatother

%% file: math_commands.tex
\usepackage{amsmath,amsfonts,bm}

\def\figref#1{figure~\ref{#1}}
\def\Figref#1{Figure~\ref{#1}}

\def\eqref#1{equation~\ref{#1}}

\def\1{\bm{1}}

\def\vx{{\bm{x}}}

\def\mW{{\bm{W}}}

\DeclareMathAlphabet{\mathsfit}{\encodingdefault}{\sfdefault}{m}{sl}
\SetMathAlphabet{\mathsfit}{bold}{\encodingdefault}{\sfdefault}{bx}{n}

\newcommand{\R}{\mathbb{R}}

\DeclareMathOperator*{\argmax}{arg\,max}